\PassOptionsToPackage{table}{xcolor}
\documentclass[11pt]{article}

\usepackage[final]{acl}

\usepackage{times}
\usepackage{latexsym}

\usepackage[T1]{fontenc}

\usepackage[utf8]{inputenc}

\usepackage{microtype}

\usepackage{inconsolata}

\usepackage{graphicx}
\usepackage{amsmath}

\usepackage{booktabs}
\usepackage{tcolorbox}
\usepackage{amssymb}
\usepackage[table]{xcolor}
\usepackage{mathtools}
\usepackage{amsthm}

\usepackage{microtype}
\usepackage{graphicx}
\usepackage{subcaption}
\usepackage{bbding}
\usepackage{makecell}
\usepackage{graphicx}
\usepackage[table]{xcolor}
\usepackage{array}
\usepackage{algorithmic}
\usepackage{algorithm}
\usepackage{enumitem}
\usepackage{fvextra}
\tcbuselibrary{breakable, skins}
\title{EpiPersona: Persona Projection and Episode Coupling for Pluralistic Preference Modeling}

\author{
Yujie Zhang \and
Weikang Yuan \and
Zhuoren Jiang$^*$ \and
Pengwei Yan \\[4pt]
Department of Information Resources Management, \\
Zhejiang University \\
\texttt{\{yj.zhang, yuanwk, jiangzhuoren, yanpw\}@zju.edu.cn} \\[2pt]
$^*$ Corresponding author
}

\begin{document}
\maketitle
\begin{abstract}
Pluralistic alignment is essential for adapting large language models (LLMs) to the diverse preferences of individuals and minority groups. However, existing personalized reward models often mix stable personal traits with episode-specific factors, limiting their ability to generalize across episodes. To address this challenge, we propose \textbf{EpiPersona}, a framework for explicit \textbf{persona–episode coupling}.
EpiPersona first projects noisy preference feedback into a low-dimensional persona space, where similar personas are aggregated into shared discrete codewords. This process separates enduring personal characteristics from situational signals without relying on predefined preference dimensions. The inferred persona representation is then coupled with the current episode, enabling episode-aware preference prediction.
Extensive experiments show that EpiPersona outperforms the baselines. It achieves notable performance gains in hard episodic-shift scenarios, while remaining effective with sparse preference data.
\end{abstract}

\section{Introduction}

Aligning artificial intelligence with human preferences has been a crucial point of both theoretical and applied research for AI systems \cite{kirk2024benefits, zhang2024personalization, zhupersonality, guan2025survey, xu2025personalized}. However, most existing approaches treat preferences as uniform, overlooking the heterogeneity found across individuals and groups \cite{li20251, oh-etal-2025-comparison}. Addressing the needs of a diverse population is therefore essential, making pluralistic alignment \cite{poddar2024personalizing, chen2025pal, ryan-etal-2025-synthesizeme} a vital step toward building human-centered AI systems \cite{zheng2023judging, zhang-etal-2025-bridging, kirk2024prism, yuan-etal-2026-lecode}.


Large language models can learn from human feedback \cite{christiano2017deep, bai2022training, xie2024can, crosshypothetical, riemerposition}. Since keeping a human in the loop to provide continuous feedback is impractical \cite{kaufmann2025survey, cai2026one}, there is a growing demand for personalized reward models \cite{ryan-etal-2025-synthesizeme, boselore} which can capture individual preferences using a small amount of user-specific annotation \cite{cai2026one, zhao2023group, chen2025pal}. 

To collect such preference feedback, studies predominantly adopt the pairwise comparison approach (e.g., ``Which response do you prefer?'') to identify personalized preferences due to its user-friendliness \cite{zheng2023judging, kirk2024prism, ryan-etal-2025-synthesizeme, balepur-etal-2025-whose, li20251, poddar2024personalizing}. Existing approaches can be categorized into \textit{non-parametric} modeling and \textit{parametric} modeling. \textit{Non-parametric} modeling leverages historical retrieval \cite{zollo2024personalllm} or text-based preference summarization \cite{balepur-etal-2025-whose, ryan-etal-2025-synthesizeme} to infer the personalized preference. While intuitive, this approach is less effective at higher-level abstractions as preferences are particularly implicit and complex \cite{jang2023personalized, balepur-etal-2025-whose}.
\textit{Parametric} modeling compresses long-term interaction histories into a semantic space, such as predefined preference dimensions \cite{oh-etal-2025-comparison, li20251} (e.g., harmlessness, helpfulness, and humor) or context compression \cite{zhao2023group, poddar2024personalizing} for downstream tasks such as pluralistic reward learning \citep{bradley1952rank}. However, the unique challenges remain unsolved. Two interconnected challenges are highlighted:

\begin{figure*}[t]
  \begin{center}
  \centerline{\includegraphics[width=0.75\textwidth]{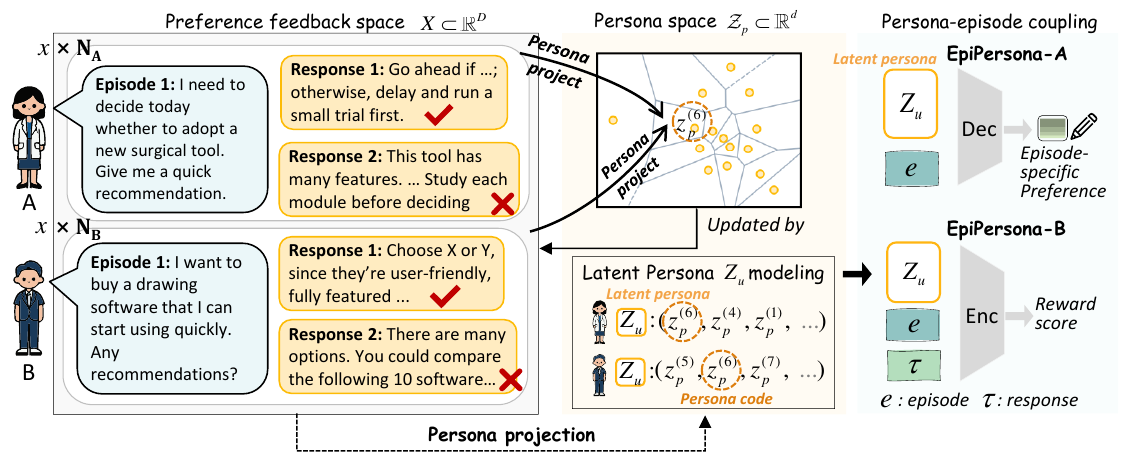}}
  \caption{We propose \textbf{EpiPersona}. Based on an individual’s historical preference information, we model the individual’s latent persona $Z_u$. We further introduce two variants of EpiPersona that couple persona $Z_u$ with episode context $e$ for preference prediction.}
  \label{fig:epipersona}
\end{center}
\end{figure*}






\textit{First}, existing methods mix stable personal traits with temporary episodic signals in the representation, failing to separate \textit{stable} traits from \textit{dynamic} episodic influences. As a result, static traits are not disentangled, limiting cross-episode generalization.
\textit{Second}, prevailing methods rely on predefined dimensions to model preferences. Human preference is better understood as a latent semantic space to be discovered \cite{poddar2024personalizing}, while predefined dimensions constrain its richness and introduce priori biases.






To address these challenges, we propose \textbf{persona projection}, which disentangles stable traits (i.e., the persona) from episode-specific preference feedback.
It is motivated by the insight that individuals' preferences are driven by their latent and stable personas. For example (see Figure \ref{fig:epipersona}), a user \textit{A} in a \textit{work decision} scenario and user \textit{B} in an \textit{online shopping} scenario may both prefer concise and direct responses; their feedback thus reflects the same \textit{efficiency-driven} persona. Formally, our method projects individuals’ preference feedback from the high-dimensional token space $X \subset \mathbb{R}^D$ into a low-dimensional \textbf{persona space} $\mathcal{Z}_p \subset \mathbb{R}^d$ ($d \ll D$), enabling feedback $x\in X$ with similar latent personas to be aggregated into shared persona codewords $\mathbf{z}_p\in \mathcal{Z}_p$. In other words, we learn a mapping $x \mapsto \mathbf{z}_p$. This mapping function converts the originally redundant and noisy feedback into a combinatorial persona codeword, enabling robust generalization across individuals.

Building on persona projection, we propose EpiPersona, a method for \textbf{PERSONA}–\textbf{EPI}sode coupling. It constructs persona codewords automatically without relying on predefined dimensions by proposing \textit{parameterized abductive reasoning} and \textit{VQ-based mapping} \cite{van2017neural}. After disentangling personas, we further \textbf{couple} them with the current episode, capturing both stable persona and situational factors, enabling robust prediction of cross-episode
preferences. Our main contributions are as follows:




$\bullet$ We propose EpiPersona, a framework that separates stable persona traits from episode-specific signals without relying on predefined preference dimensions.

$\bullet$ We propose parameterized abductive reasoning and VQ-based persona projection to ground preference feedback into a discrete persona codebook, coupled with episode context for contextualized preference prediction.

$\bullet$ Extensive experiments demonstrate that EpiPersona outperforms baselines, with particularly robust generalization in challenging settings such as the episodic-shift scenario.





\section{Related work}



Pluralistic alignment adapts the preferences of diverse groups, especially minority populations, thereby maximizing the collective benefits \cite{zollo2024personalllm} and promoting fairness across heterogeneous user communities \cite{kirk2024prism, kirk2024benefits}. Studies show that even high-performing models struggle to satisfy preferences across different regions, cultures, and topics \cite{kirk2024prism}. These challenges underscore the diversity and complexity of human preferences. Due to the sparseness of individual preference data \cite{oh-etal-2025-comparison, li20251}, existing approaches use human feedback in the form of pairwise preference comparisons \cite{zheng2023judging, poddar2024personalizing, ryan-etal-2025-synthesizeme, balepur-etal-2025-whose, li20251, kirk2024prism}. Prior personalized reward models employ the following methods to model pluralistic preferences:

\textbf{Non-parametric modeling.}
\emph{Preference summarization (text-based).} Preference dimensions are often defined as such attributes as helpfulness, honesty, and harmlessness \cite{Askell2021AGL}. However, simplified dimensions may not be able to capture the diversity of human preferences \cite{kirk2024benefits}. Studies \cite{ryan-etal-2025-synthesizeme, balepur-etal-2025-whose} utilize \textit{preference attribution} methods to extract individual preferences. Among them, SynthesizeMe \cite{ryan-etal-2025-synthesizeme} uses bootstrap reasoning on limited user interaction data and verifies inferred preferences on the validation set to synthesize the personas. PI \cite{balepur-etal-2025-whose} leverages \textit{text-based} abductive reasoning on preference data to infer users’ underlying needs and interests. They can be adaptive to in-context learning, such as \textit{LLM-as-a-judge} optimizing \cite{gu2024survey} and personal LLM \cite{ryan-etal-2025-synthesizeme, li20251, balepur-etal-2025-whose, du2025valuesim}.
\emph{Retrieval-augmented generation (RAG).} PersonalLLM \cite{zollo2024personalllm} combines context learning with historical information retrieval to produce the personalized responses.
In contrast, EpiPersona employs the parameterized modeling approach to globally and abstractly represent personalized preferences.







\textbf{Parametric modeling.}
Parameterized preferences have been used for personalized reward model training \cite{zhao2023group, poddar2024personalizing}. For example, GPO \cite{zhao2023group} adds a few-shot preference module on top of the base LLM and learns to predict group preference scores from historical user data. VPL \cite{poddar2024personalizing} compresses historical user preference data into a parameterized latent distribution using the VAE module. 
\emph{Predefined preference dimensions.} AMPLe \cite{oh-etal-2025-comparison} models user preferences along multiple predefined dimensions via posterior updates, while \citet{li20251} constructs a preference space grounded in psychological models. In contrast, EpiPersona requires no predefined dimensions, instead disentangling stable personas directly from episode-specific feedback.

\section{Problem Setting}

Let $\mathcal{U}$ denote a population of individuals and $\mathcal{E}$ the space of episodes. 
Each individual $u \in \mathcal{U}$ may participate in multiple episodes $e \in \mathcal{E}$, 
where each episode is associated with a query $q_e$. 
For a given episode $e$, the system presents a set of candidate responses 
$\{\tau_{e}^{(i)}\}_{i=1}^{K}$. 
We assume that preferences are generated by an underlying, unobserved reward function:
\begin{equation}
R_u(\tau_{e})  \rightarrow \mathbb{R},
\end{equation}
which reflects the utility assigned to response $\tau_{e}$ by individual $u$ under episode $e$.

\paragraph{Observed feedback.}
The reward function $R_u$ is not directly observable.
Instead, for each individual $u$ and episode $e$, we observe the pairwise preference feedback of the form:
\begin{equation}
x = (\tau_{e}^{(i)}, \tau_{e}^{(j)}, y_e),
\end{equation}
where $y_e \in \{0,1\}$. $y_e =1 $ indicates $\tau_{e}^{(i)}$ is preferred over $\tau_{e}^{(j)}$ in episode $e$.
We denote the collection of observations for individual $u$ as:
\begin{equation}
X_u = \{ (\tau_{e}^{(i)}, \tau_{e}^{(j)}, y_e) \mid e \in \mathcal{E}_u \}.
\end{equation}

\paragraph{Latent persona assumption.}

We define the episode-related response $\tau_e$
and a latent persona $Z_u$ of individual $u$.
We assume that the underlying reward function can be modeled as:
\begin{equation}
R_u(\tau_e)  \;\hat{=}\; R_{\theta}(Z_u, \tau_e) \in \mathbb{R},
\end{equation}
where \(R_{{\theta}}(\cdot, \cdot)\) is an estimated reward function. The latent persona $Z_u$ is unobserved but can be inferred from observable features $X_u$:
\begin{equation}
Z_u \sim p(Z_u \mid X_u).
\end{equation}

\section{Method}
We present an overview of EpiPersona in Figure \ref{fig:epipersona}. First, we disentangle individuals’ stable persona traits from their preference feedback via persona projection, mapping feedback with similar personas into a shared latent space (Section \ref{sec:projection}). Subsequently, the model couples the inferred latent persona with the specific episodes to enable contextualized preference prediction (Section \ref{sec:coupling}).
The EpiPersona achieves persona disentanglement through parameterized abductive reasoning (Section \ref{sec:abd}) and vector-quantized mapping (Section \ref{sec:vq}), automatically constructing a structured persona space without requiring predefined preference dimensions.

\subsection{Persona projection}
\label{sec:projection}
\paragraph{Formulation and assumptions.} We assume that each observed preference feedback $x \in X_u$ from individual $u$ is jointly determined by the episode-specific responses $\tau_e$ and a latent, stable persona $Z_u$. Formally, the preference formation process can be expressed as:

\begin{equation}
y_e \sim p\left(y_e \mid Z_u, \tau_e^{(i)}, \tau_e^{(j)}\right).
\end{equation}

\begin{figure}[ht]
  \begin{center}  
     \centerline{\includegraphics[width=0.7\columnwidth]{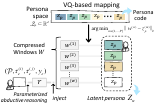}}
    \caption{Overview of the persona projection mapping.}
    \label{fig:detail_small_model}
  \end{center}
\end{figure}

\paragraph{Latent persona modeling.}  
The overall persona projection mapping function is illustrated in Figure~\ref{fig:detail_small_model}.
Specifically, we model an individual’s persona $Z_u$ as a latent variable that potentially follows a \textit{multimodal} distribution, reflecting that a single individual may exhibit multiple persona patterns \cite{chen2025pal}. Concretely, we represent the latent persona of individual $u$ as a set of persona codewords:
\begin{equation}
Z_u = \{\, z_p \mid z_p \in \mathcal{Z}_p \,\}
\end{equation}
where each $z_p$ corresponds to a stable persona codeword. Crucially, each persona codeword $z_p$ lies in a \emph{shared persona space} $\mathcal{Z}_p$, which is common to all individuals. To further represent the persona codeword ${z}_p$, we leverage the individual's feedback $x\in X_u$. Specifically, we map feedback instance $x$ into the persona codeword $z_p\in \mathcal{Z}_p$ via a persona projection function $g$:
\begin{equation}
g: x \mapsto z_p, \quad z_p \in \mathcal{Z}_p \subset \mathbb{R}^d,
\label{eq:g}
\end{equation}
where $d \ll D$ and $D$ is the dimension of the original feedback $x \in \mathbb{R}^D$. It learns to automatically construct the persona space. 

In this way, $Z_u = \{\, g(x) \mid x \in X_u \,\} \subseteq \mathcal{Z}_p$ captures the diverse persona traits of an individual, while grounding them in the shared persona space $\mathcal{Z}_p$. This shared space $\mathcal{Z}_p$ is also jointly updated from feedback across all individuals, enabling it to retain collective universality.






\paragraph{Objective of persona projection.} In summary, the objective of persona projection is to map episode-specific feedback $x$ into a low-dimensional persona codeword $z_p\in \mathcal{Z}_u$. This process aims to disentangle stable persona from episodic variability. We realize the mapping $g$ through \textit{parameterized abductive reasoning} (Section~\ref{sec:abd}) and \textit{VQ-based mapping} (Section~\ref{sec:vq}).

\subsection{Persona-episode coupling}
\label{sec:coupling}


Based on persona projection, we couple the user persona $Z_u$ with the episode to model an individual's preferences. This process captures how an individual's persona may be activated in an episode. 

Since downstream scenarios differ fundamentally in output space, we propose 
two variants: \textit{generative} for preference generation (natural language), 
and \textit{discriminative} for preference scoring.

\textbf{EpiPersona-A (\textit{generative}):} This method generates a natural language description of the individual preference under the episode $e$. Note that the input only includes the latent persona $Z_u$ (based on historical feedback) and episode $e$. It allows the generated $\text{Pref}_e$ to be used in a plug-and-play manner, such as judging the individuals' potential choices.
Formally, the process is defined as:
\begin{equation}
\text{Pref}_e = g_\psi(Z_u, e)    
\end{equation}
\noindent where $g_\psi$ is a transformer-based language model trained with an NTP \cite{radford2018improving} objective. The output follows a JSON format containing keys such as \textit{persona}, \textit{value}, \textit{identification}, \textit{intent}, \textit{style}, etc. Detailed comparison of EpiPersona-A with related methods is in Appendix \ref{app:acompare}.

\textbf{EpiPersona-B (\textit{discriminative}):} The reward is modeled via a Bradley-Terry preference comparison \cite{bradley1952rank} based on latent persona $Z_u$:
\begin{equation}
\label{equ:rewardmodel}
R_{\theta}(Z_u, \tau_e) = \mathbf{W}^{\top} \phi\bigl(h_{\theta_h}(Z_u, \tau_e)\bigr)    
\end{equation}
where $h_{\theta_h}$ is an encoder producing episode-persona coupled representation,
    $\phi$ is an optional nonlinear transformation,
    $\mathbf{W}$ maps the representation to a scalar reward.
More detailed model architecture is shown as Figure \ref{fig:detail_epipersona} in the Appendix. For the inference algorithm, see the Alg.\ref{alg:epipersona-a} and \ref{alg:epipersona-b-mod} in the Appendix.

\subsection{Parameterized abductive reasoning}
\label{sec:abd}
To realize the persona projection function $g$ (see Eq.\ref{eq:g}), we first implement the instance-level encoding, mapping each preference-feedback instance to an estimated persona $r$.

We approach this using \emph{abductive reasoning} following recent work \cite{balepur-etal-2025-whose, ryan-etal-2025-synthesizeme}, which infers a plausible persona from observed data. Building on this idea, we introduce \emph{parameterized} abductive reasoning, which estimates latent preference by comparing choices. Specifically, it does not require the LLM to generate a human-readable reasoning trace. The \textit{estimated persona} $r$ is represented as a continuous learnable vector:

\begin{equation}
\lambda(\mathcal{P}, \tau_e^{(i)}, \tau_e^{(j)}, y_e) \mapsto \mathit{r},
\end{equation}
where $\mathcal{P}$ denotes the prompt that elicits the model to infer the persona underlying the user’s choice, $\lambda$ is the LLM-based encoder. The estimated persona $r$ encodes the rationale behind an individual's preference feedback. At this stage, $r$ is still instance-level and \emph{not} shared across users.

\subsection{VQ-based mapping}
\label{sec:vq}
Building on parameterized abductive reasoning (see Section \ref{sec:abd}), we then map the estimated persona $r$ into a shared persona space using a VQ-based mapping, completing the persona projection function $g$. 
The procedure is as follows:

\paragraph{Variable persona compressed window.}
Let the estimated persona $\mathbf{r}\in\mathbb{R}^d$ be injected into compressed windows as an embedding.
The resulting window representation is then projected into a shared representation space via $f_{\theta_f}(\cdot)$, yielding $\mathbf{w}^{(w)}=f_{\theta_f}(r).$
Each window $\mathbf{w}^{(w)}$ thus encapsulates a latent estimated persona derived from observed individual feedback. The window size is flexibly configurable.


\paragraph{Persona codebook.}  
We then define the persona codebook $\mathcal{Z}_p$, which represents the shared persona space. $P$ denotes the codebook size. Formally, $\mathcal{Z}_p$ is defined as:
\begin{equation}
\mathcal{Z}_p = \{ z_p^{(1)}, z_p^{(2)}, \dots, z_p^{(P)} \}, \quad z_p \in \mathbb{R}^{d}.
\end{equation}

VQ-based mapping aims to map the compressed windows onto a set of learnable persona codewords $z_p$, without assuming the predefined dimensions.

\paragraph{Nearest-neighbor quantization.}  
We represent the $n$ windows of user $u$ as
$
\mathcal{W} = [\mathbf{w}_1,\ldots,\mathbf{w}_n]^\top \in \mathbb{R}^{n\times d}.
$
$\mathbf{w}^{(w)}$ is quantized by selecting the nearest persona codeword from $\mathcal{Z}_p$:  
\begin{equation}
m^{*(w)} = \arg\min_{m \in \{1,\dots,P\}} \| \mathbf{w}^{(w)} - z_p^{(m)} \|_2.
\end{equation}  
The corresponding persona codeword is then defined as: 
\begin{equation}
\mathbf{Z}_u = \bigl( z_p^{(m^{*(w)})} \bigr)_{w=1}^{n}.
\end{equation}

\paragraph{Loss.}
To train the VQ module, we adopt two auxiliary objectives: a commitment loss $\mathcal{L}_{\text{commit}}$ and an entropy-based diversity loss $\mathcal{L}_{\text{div}}$. 

\textbf{Diversity regularisation.}
To avoid codebook collapse \cite{Zhu2024AddressingRC}, we propose an entropy-based diversity loss. For each window representation $\mathbf{w}^{(w)}$, we compute a soft assignment over the codebook via cosine similarity with temperature $\tau$:    
\begin{equation}
a_c^{(w)} = \frac{
  \exp\!\bigl(\mathrm{cos}(\mathbf{w}^{(w)},\,z_p^{(c)}) \,/\, \tau\bigr)
}{
  \sum_{c'=1}^{P} \exp\!\bigl(\mathrm{cos}(\mathbf{w}^{(w)},\,z_p^{(c')}) \,/\, \tau\bigr)
}.
\end{equation}
We average the soft assignments across all $n$ windows to obtain the empirical codeword utilisation distribution $\bar{p}_c = \frac{1}{n}\sum_{w=1}^{n} a_c^{(w)}$, and define the diversity loss as its negative entropy:
\begin{equation}
\mathcal{L}_{\text{div}} = -H(\bar{\mathbf{p}}) = \sum_{c=1}^{P} \bar{p}_c \log \bar{p}_c \;\leq\; 0.
\end{equation}
Minimising $\mathcal{L}_{\text{div}}$ is equivalent to maximising $H(\bar{\mathbf{p}})$, thereby encouraging uniform utilisation across all $P$ persona codewords. Crucially, the gradient flows through   
$\mathbf{w}^{(w)}$ back to the encoder $f_{\theta_f}$, while the codebook is detached, so only the encoder weights are steered toward producing more diverse representations.


\textbf{Commitment loss.} $\mathcal{L}_{\text{commit}}$ is a standard VQ commitment \cite{van2017neural} loss that encourages the projected window representations to stay close to their assigned codebook entries. It prevents the encoder from producing representations that drift arbitrarily between codewords:
\begin{equation}
\mathcal{L}_{\text{commit}} = \sum_{w=1}^{n} \bigl\| \mathbf{w}^{(w)} - \mathrm{sg}\bigl[z_p^{(m^{*(w)})}\bigr] \bigr\|_2^2,
\end{equation}
where $\mathrm{sg}[\cdot]$ denotes the stop-gradient operator. Gradients are propagated back to the encoder via the straight-through estimator \cite{bengio2013estimating}, while the codebook entries are updated through an exponential moving average (EMA). The details of EMA are provided in Appendix~\ref{app:ema_commit}.

\subsection{Overall training objective}
The overall training objective combines the task loss $\mathcal{L}_{\text{task}}$ with the two VQ auxiliary terms:
\begin{equation}
\mathcal{L} = \mathcal{L}_{\text{task}} + \alpha_{\text{commit}}\,\mathcal{L}_{\text{commit}} + \alpha_{\text{div}}\,\mathcal{L}_{\text{div}},
\end{equation}
where $\alpha_{\text{commit}}$ and $\alpha_{\text{div}}$ are the corresponding weights.

Corresponding to Section \ref{sec:coupling}, the task loss of EpiPersona-A is the autoregressive next-token prediction loss:
\begin{equation}
\mathcal{L}_{\text{task}}^A = -\sum_{t=1}^T \log p_{\psi}(s_t \mid s_{<t}, Z_u, e),
\end{equation}

where $s = (s_1, \ldots, s_T)$, which is the episode-specific preference description.

The task loss of EpiPersona-B is the Bradley-Terry pairwise comparison loss:
\begin{equation}
\Delta R_e = R_\theta(Z_u, \tau_e^{(i)}) - R_\theta(Z_u, \tau_e^{(j)}),
\end{equation}
then:
\begin{equation}
\mathcal{L}_{\text{task}}^B = -y_e \log \sigma(\Delta R_e) - (1 - y_e) \log(1 - \sigma(\Delta R_e)).
\end{equation}
Together, these objectives jointly optimize the task performance and update VQ representations.

\section{Experiment}
\subsection{Experiment setting}
\paragraph{Baselines.}
We compare EpiPersona against two categories of methods:
\textit{\textbf{a) Non-parametric preference modeling.}} These methods model preferences using natural language and can be adapted to LLMs in a plug-and-play manner: \textbf{SynthesizeMe}~\cite{ryan-etal-2025-synthesizeme} synthesizes personas through LLM-based bootstrap reasoning and validation; \textbf{PersonalLLM}~\cite{zollo2024personalllm} retrieves similar interactions as in-context examples to adapt to individual preferences; \textbf{PI}~\cite{balepur-etal-2025-whose} infers personas from preference feedback and incorporates them into context for alignment.
\textit{\textbf{b) Parametric preference modeling.}} \textbf{GPO}~\cite{zhao2023group} uses a transformer to predict preferences conditioned on context samples; \textbf{VPL}~\cite{poddar2024personalizing} employs a VAE to model individual preferences as latent distributions; \textbf{Bradley-Terry (BT)}~\cite{bradley1952rank} learns reward scores from pairwise comparisons without personalization. \textbf{LoRe}~\cite{boselore} represents individual reward functions as weighted combinations of shared basis functions in a low-rank subspace; \textbf{MRM}~\cite{cai2026one} reformulates personalized reward modeling as meta-learning via a MAML-style framework.

\paragraph{Tasks} 
We evaluate on two tasks with different output spaces. This setting aims to assess the personalized preferences modeling ability comprehensively. Task selection is based on downstream task requirements.
\textbf{a) Generative.} Following existing work \cite{ryan-etal-2025-synthesizeme}, this task evaluates the ability to generate natural language descriptions of individual preferences. Given an individual's historical data and the current episode, the model generates the individual's preference. An LLM judge then uses it to predict which of two candidate responses the individual would prefer.
\textbf{b) Discriminative.} This task evaluates the ability to predict personalized preference scores directly. Following standard practice~\cite{zhao2023cross, cai2026one}, we evaluate on pairwise comparisons and report accuracy. See Appendix~\ref{app:task} for the complete task formulation.

\paragraph{Datasets} \textit{\textbf{Prism}} 
\cite{kirk2024prism} is a values- and controversy-guided human feedback dataset linking sociodemographics and stated preferences of 1,500 participants from 75 countries. It provides pairwise comparisons of model responses within shared conversational contexts. \textit{\textbf{Arena}} \cite{zheng2023judging} is a crowdsourced benchmark where users compare two anonymous LLM outputs and select their preferred response. The distribution and characteristics of the dataset are shown in Appendix \ref{app:dataset}.

\textbf{Dataset setting.} The datasets contains no overlapping individuals between the train and test sets. Moreover, the historical and current samples do not share overlapping conversations.
This setup ensures we evaluate on \emph{\textbf{unseen}} individuals with \emph{\textbf{unseen}} conversations, which aligns with real-world scenarios.




\paragraph{Implementation setup} 
Fine-tuning is performed using the LoRA method applied to the main linear layers. The VQ codebook is initialized using K-means. $\lambda$ is consistent with the finetuned backbone model used. Additional implementation details and hyperparameter experiments are provided in Appendix \ref{app:detail} and \ref{app:hyper}.
\textbf{Additional setup for generative setting:} For downstream evaluation (LLM-as-a-Judge), LLMs including GPT-OSS-120B and Llama-3.3-70B-Instruct are employed. The backbone model used is Llama-3.1-8B. \textbf{Additional setup for discriminative setting:} The backbone model of all baselines used is Llama-3.1-8B and Llama-3.2-3B. The source code is available at \url{https://github.com/PiLab-ZJU/EpiPersona}.

\begin{table*}[h]
\caption{Comparison on generative task (LLM-as-a-judge). Values are percentages with standard deviations (95\% CI).}
\label{table:llm-as-judge}
\centering
\small
\setlength{\tabcolsep}{4pt}
\renewcommand{\arraystretch}{0.82}
\definecolor{lightgray}{RGB}{235,235,235}

\begin{tabular}{lcccc}
\toprule
 & \multicolumn{2}{c}{\textbf{Prism}} & \multicolumn{2}{c}{\textbf{Arena}} \\
\cmidrule(r){2-3} \cmidrule(r){4-5}
 & GPT-OSS-120B & Llama-3.3-70B & GPT-OSS-120B & Llama-3.3-70B \\
\midrule
\rowcolor{lightgray}
\multicolumn{5}{l}{\textit{Baseline}} \\
Random & 50\% & 50\% & 50\% & 50\% \\
LLM as a judge (Base model) & 56.30 {\scriptsize $\pm$ 1.26\%} & 55.90 {\scriptsize $\pm$ 1.22\%} & 64.36 {\scriptsize $\pm$ 3.91\%} & 64.18 {\scriptsize $\pm$ 3.91\%} \\
\midrule
\rowcolor{lightgray}
\multicolumn{5}{l}{\textit{Personalization Methods}} \\
SynthesizeMe (GPT-OSS-120B) & 57.24 {\scriptsize $\pm$ 1.27\%} & 57.67 {\scriptsize $\pm$ 1.27\%} & \underline{64.89} {\scriptsize $\pm$ 4.11\%} & \underline{64.39} {\scriptsize $\pm$ 3.93\%} \\
SynthesizeMe (Llama-3.1-8B) & 57.27 {\scriptsize $\pm$ 1.28\%} & 56.86 {\scriptsize $\pm$ 1.26\%} & 63.83 {\scriptsize $\pm$ 4.26\%} & 64.17 {\scriptsize $\pm$ 3.85\%} \\
PI (GPT-OSS-120B) & 54.00 {\scriptsize $\pm$ 1.37\%} & 57.22 {\scriptsize $\pm$ 1.28\%} & 61.72 {\scriptsize $\pm$ 4.78\%} & 63.28 {\scriptsize $\pm$ 3.98\%} \\
PI (Llama-3.1-8B) & \underline{57.48} {\scriptsize $\pm$ 1.28\%} & 56.58 {\scriptsize $\pm$ 1.36\%} & 63.06 {\scriptsize $\pm$ 4.13\%} & 63.97 {\scriptsize $\pm$ 3.97\%} \\
PersonalLLM (demo num=1) & 54.39 {\scriptsize $\pm$ 1.26\%} & 55.81 {\scriptsize $\pm$ 1.24\%} & 59.71 {\scriptsize $\pm$ 4.10\%} & 59.33 {\scriptsize $\pm$ 4.09\%} \\
PersonalLLM (demo num=3) & 56.29 {\scriptsize $\pm$ 1.38\%} & 56.42 {\scriptsize $\pm$ 1.25\%} & 62.01 {\scriptsize $\pm$ 4.13\%} & 63.13 {\scriptsize $\pm$ 3.89\%} \\
PersonalLLM (demo num=5) & 56.46 {\scriptsize $\pm$ 1.34\%} & 56.21 {\scriptsize $\pm$ 1.24\%} & 62.43 {\scriptsize $\pm$ 4.97\%} & 64.13 {\scriptsize $\pm$ 3.92\%} \\
\midrule
\rowcolor{lightgray}
\multicolumn{5}{l}{\textit{General LLMs (Persona summarization)}} \\
Qwen3-235B & 57.09 {\scriptsize $\pm$ 1.21\%} & \underline{58.40} {\scriptsize $\pm$ 1.12\%} & 60.97 {\scriptsize $\pm$ 4.13\%} & 61.10 {\scriptsize $\pm$ 4.03\%} \\
GPT-OSS-120B & 56.98 {\scriptsize $\pm$ 1.38\%} & 57.79 {\scriptsize $\pm$ 1.27\%} & 63.04 {\scriptsize $\pm$ 4.15\%} & 64.32 {\scriptsize $\pm$ 3.93\%} \\
GPT-4o-mini & \underline{57.48} {\scriptsize $\pm$ 1.28\%} & 58.21 {\scriptsize $\pm$ 1.25\%} & 61.39 {\scriptsize $\pm$ 4.10\%} & 60.37 {\scriptsize $\pm$ 4.03\%} \\
LLAMA-3.1-8B & 56.82 {\scriptsize $\pm$ 1.64\%} & 58.39 {\scriptsize $\pm$ 1.24\%} & 61.27 {\scriptsize $\pm$ 4.15\%} & 63.99 {\scriptsize $\pm$ 4.10\%} \\
LLAMA-3.3-70B & 56.78 {\scriptsize $\pm$ 1.25\%} & 58.18 {\scriptsize $\pm$ 1.29\%} & 63.89 {\scriptsize $\pm$ 3.97\%} & 62.68 {\scriptsize $\pm$ 4.00\%} \\
\midrule
\rowcolor{lightgray}
\multicolumn{5}{l}{\textit{Ours}} \\
\textbf{EpiPersona-A} & \textbf{59.38} {\scriptsize $\pm$ 1.25\%} & \textbf{59.15} {\scriptsize $\pm$ 1.30\%} & \textbf{66.07} {\scriptsize $\pm$ 3.98\%} & \textbf{65.01} {\scriptsize $\pm$ 3.90\%} \\
\bottomrule
\end{tabular}
\end{table*}

\subsection{Results}
Experiments and analyses are presented in Sections~\ref{sec:performance} and~\ref{sec:analysis}, respectively.

\subsubsection{Performance validation}
\label{sec:performance}
We evaluate \textit{EpiPersona} from two aspects: 1) Performance validation, 2) Contribution of each component to preference modeling.

\paragraph{Generative setting.}





As shown in Table~\ref{table:llm-as-judge}, we compare EpiPersona-A with non-parametric preference modeling approaches. Against the RAG-based \textit{PersonalLLM}, our method improves by ~$2.92\%$ on \textit{Prism} and ~$3.64\%$ on \textit{Arena}. While RAG-based approaches capture localized or recent information, they overlook globally stable preferences. Conversely, methods like \textit{SynthesizeMe} and \textit{PI} model static global personas but struggle to adapt to episode-dependent preference variations. EpiPersona-A addresses both limitations by conditioning episode-specific preference predictions on a latent persona, with parametric history modeling enabling better generalization and robustness.

\paragraph{Discriminative setting.} As shown in Table~\ref{table:reward_model}, we compare EpiPersona-B (ours) with parametric preference modeling methods on pluralistic reward learning. EpiPersona-B consistently achieves higher or comparable performance than other baselines, demonstrating the effectiveness of disentangling the persona from episode-related information via persona projection. Specifically, while \textit{VPL} and \textit{MRM} model latent individual information, their performance is sensitive to context and model size. In contrast, EpiPersona-B shows robust preference prediction, achieving the highest scores with both Llama-3.2-3B and Llama-3.1-8B.

\begin{table}[h]
\caption{Comparison of the discriminative task. Values are percentages with standard deviations (95\% CI).}
\label{table:reward_model}
\centering
\scalebox{0.68}{\begin{tabular}{lcccc}
\toprule
 & \multicolumn{2}{c}{\textbf{Prism}} & \multicolumn{2}{c}{\textbf{Arena}} \\
\cmidrule(r){2-3} \cmidrule(r){4-5}
 & Llama-3.2-3B & Llama-3.1-8B & Llama-3.2-3B & Llama-3.1-8B \\
\midrule
GPO & 55.26 {\scriptsize $\pm$ 1.25\%} & 56.48 {\scriptsize $\pm$ 1.24\%} & 53.87 {\scriptsize $\pm$ 4.14\%} & \underline{56.69} {\scriptsize $\pm$ 4.05\%} \\
VPL & \underline{58.26} {\scriptsize $\pm$ 1.75\%} & \underline{58.23} {\scriptsize $\pm$ 1.75\%} & 56.69 {\scriptsize $\pm$ 5.81\%} & 54.93 {\scriptsize $\pm$ 5.71\%} \\
BT & 56.58 {\scriptsize $\pm$ 1.47\%} & 57.11 {\scriptsize $\pm$ 1.55\%} & 51.94 {\scriptsize $\pm$ 5.31\%} & 54.06 {\scriptsize $\pm$ 5.42\%} \\
LoRe & 53.57  {\scriptsize $\pm$ 1.81\%}  & 51.70 {\scriptsize $\pm$ 1.78\%}  & 53.17 {\scriptsize $\pm$ 5.99\%}  &  52.48 {\scriptsize $\pm$ 5.63\%} \\
MRM & 55.65 {\scriptsize $\pm$ 1.75\%}  & 57.14 {\scriptsize $\pm$ 1.71\%}  &  \underline{56.79} {\scriptsize $\pm$ 5.63\%} & 56.68 {\scriptsize $\pm$ 5.63\%}  \\
\midrule
Ours & \textbf{60.23 {\scriptsize $\pm$ 1.76\%}} & \textbf{58.54 {\scriptsize $\pm$ 1.76\%}} & \textbf{59.57} {\scriptsize $\pm$ 5.67\%} & \textbf{57.45 {\scriptsize $\pm$ 5.12\%}} \\
\bottomrule
\end{tabular}}
\end{table}




\paragraph{Ablation of EpiPersona.} 
To validate the contribution of each component in EpiPersona, we conduct ablation studies on two datasets, focusing on a more complex generative setting. Table~\ref{table:ablation} presents the results comparing two main ablation variants. Additionally, a detailed ablation study on the effect of the diversity loss on codebook utilization and training stability is provided in Appendix~\ref{app:diversity_ablation}.


\textbf{Ablation variant 1: Text-based preference modeling} (w/o modeling preference on latent persona). \textit{Global persona prediction} summarizes the text-based persona from the historical information. \textit{Episode-specific preference prediction} couples the text-based persona with the episode to predict the episode-specific preference (see prompt in Appendix \ref{box:guidance}.4).

\textbf{Ablation variant 2: Parametric preference modeling} represents individual preferences with a latent persona $Z_u$. To assess the contribution of specific components, we conducted ablation studies by removing the VQ-based mapping (\textit{w/o VQ-based mapping}) and parametric abductive reasoning (\textit{w/o param. abductive reasoning}) modules.

Component-wise analysis. For \textbf{ablation variant 1 (Text-based preference modeling)}, using the stable global persona achieves moderate performance.
Adding episode-specific preference prediction with different LLMs brings minor improvements, suggesting that \textit{text-based} approaches alone have limited ability to capture episode-level preferences.


For \textbf{ablation variant 2 (Parametric preference modeling)}, representing individual preferences with a latent persona $Z_u$ boosts performance. Ablation studies show that removing the VQ-based mapping or the parametric abductive reasoning module reduces accuracy, confirming that each component contributes to overall performance.


\begin{table}[h]
\caption{Ablation study on EpiPersona-A.}
\label{table:ablation}
\centering
\scalebox{0.7}{
\begin{tabular}{lcc}
\toprule
 & \textbf{Prism} & \textbf{Arena} \\
\midrule
\rowcolor{cyan!10} \multicolumn{3}{l}{Ablation Variant 1: \textbf{Text-based} preference modeling} \\
\underline{\textit{Global persona prediction}} & & \\
only stable persona gpt-oss-120b & 57.54 {\scriptsize $\pm$ 1.15\%} & 64.05 {\scriptsize $\pm$ 4.13\%} \\
\underline{\textit{Episode-specific preference prediction}} & & \\
with Qwen3-235B & 57.09 {\scriptsize $\pm$ 1.21\%} & 60.97 {\scriptsize $\pm$ 4.13\%} \\
with distilled Llama-3.1-8B & 56.98 {\scriptsize $\pm$ 1.38\%} & 62.52 {\scriptsize $\pm$ 4.08\%} \\
\midrule
\rowcolor{cyan!10} \multicolumn{3}{l}{Ablation Variant 2: \textbf{Parameter-based} preference modeling} \\
w/o VQ-based mapping & 57.56 {\scriptsize $\pm$ 1.13\%} & 62.66 {\scriptsize $\pm$ 4.01\%} \\
w/o param. abductive reasoning & 57.99 {\scriptsize $\pm$ 1.13\%} & 64.36 {\scriptsize $\pm$ 4.01\%} \\
EpiPersona (Ours) & \textbf{59.38 {\scriptsize $\pm$ 1.25\%}} & \textbf{66.07 {\scriptsize $\pm$ 3.98\%}} \\
\bottomrule
\end{tabular}}
\end{table}

\begin{figure}[ht]
  \begin{center}
    \raggedleft
    \centerline{\includegraphics[width=1.05\columnwidth]{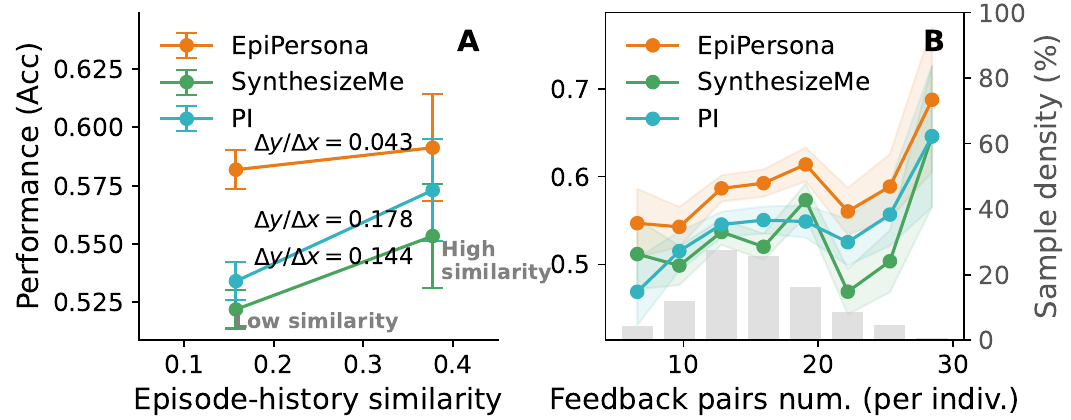}}
    \caption{Evaluation of EpiPersona under different scenarios.
(A) \textit{Episode similarity}: performance under high vs. low episode similarity. 
(B) \textit{Number of preference feedback instances}: model performance across users with varying amounts of historical feedback. 
}
    \label{fig:analysis}
  \end{center}
\end{figure}


\subsubsection{Analysis}
\label{sec:analysis}
To understand episode generalization, we analyze it from three perspectives:
(1) \textit{episode similarity},
(2) the \textit{number of individual preference feedback instances}, \textit{i.e.}, 
reflecting the data distribution, and
(3) insights from the codebook.


\paragraph{Analysis of episode similarity effects.} \emph{Episode similarity} refers to the similarity between an individual's current episode and their historical preference episodes.
Low similarity indicates a large episodic gap and shift (i.e., an \emph{episode-shift scenario}), implying that historical episodes are less informative to the current sample. Episode similarity is computed using \textit{Qwen3-Embedding}.
As shown in Figure \ref{fig:analysis}(A), the results reveal two interesting findings. \textit{\textbf{Finding 1}}: EpiPersona, \textit{SynthesizeMe}, and \textit{PI} all perform better under high episode similarity, suggesting that similar contexts help models more accurately infer individual preferences. This also indicates that samples in the \emph{episode-shift scenario} are relatively hard. \textit{\textbf{Finding 2}}: Under low episode similarity, our method shows a smaller performance drop, demonstrating its ability to \emph{handle unseen episodes} and improve generalization during inference: our model drops only $0.043$, compared to $0.178$ and $0.144$ for the other methods. In the low-episode similarity scenario, our method outperforms the SOTA by $\sim$5\% (see Figure \ref{fig:analysis}(A)).

\paragraph{Analysis of instance number.} As shown in Figure \ref{fig:analysis}(B), across varying ranges of user preference counts, our method maintains consistently competitive performance compared to the baselines. The benefits of EpiPersona are especially apparent when user preferences are sparse.

These two analyses collectively validate EpiPersona's strong generalization capability: it maintains robust performance under both episodic distribution shift and data scarcity, outperforming SOTA baselines in both challenging settings.



\paragraph{Analysis on codebook}

\begin{figure}[ht]
  \begin{center}
\centerline{\includegraphics[width=1.05\columnwidth]{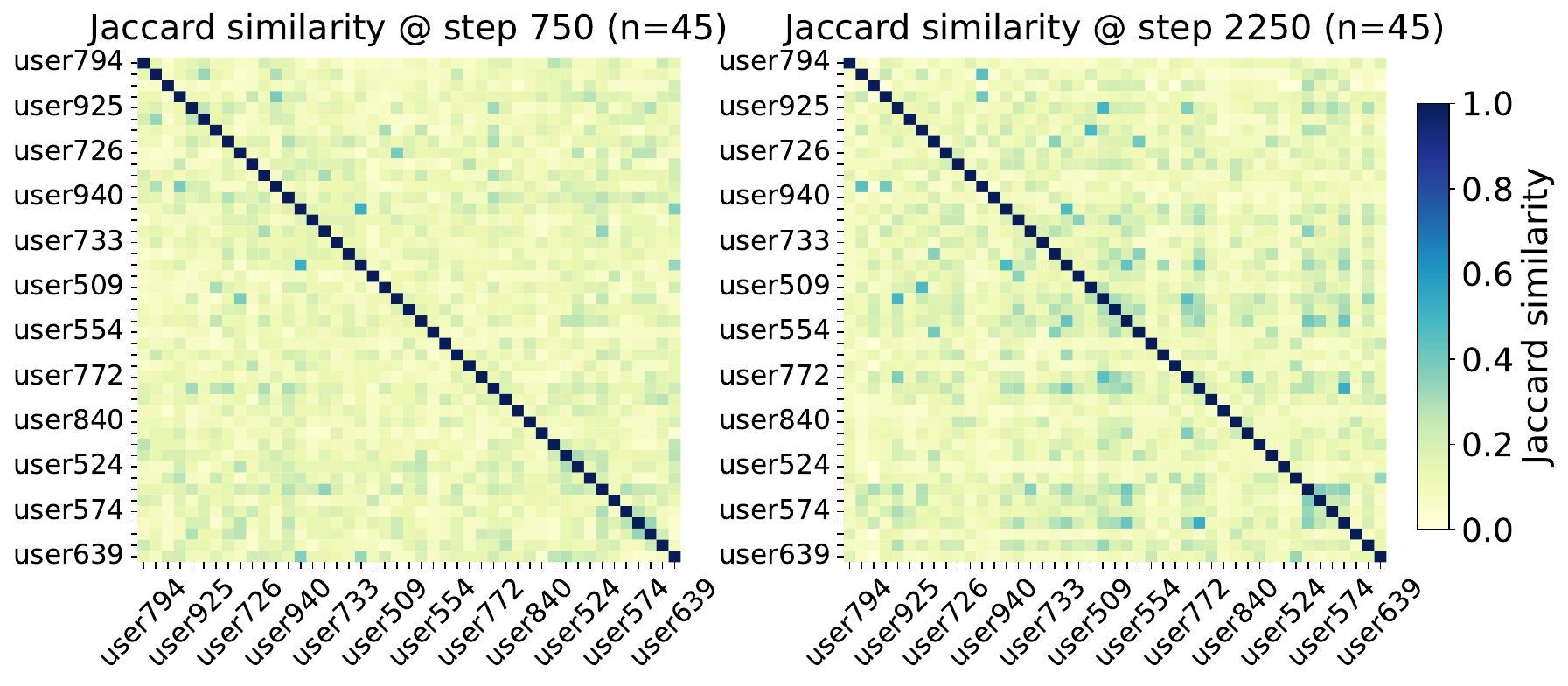}}
    \caption{Pairwise Jaccard similarity of codebook index sets among 45 randomly sampled users.}
    \label{fig:indices}
  \end{center}
\end{figure}

\textit{Codebook evolution}. Figure~\ref{fig:indices} visualizes the pairwise Jaccard similarity of codebook index sets among 45 randomly sampled users across training steps. It shows that: $\bullet$  \textit{Step 750}: Off-diagonal entries are predominantly near zero, indicating largely distinct index sets across users.
$\bullet$ \textit{Step 2250}: Partial clusters of similar users emerge, suggesting the codebook begins to organize users by shared preferences.

It indicates that the learned codebook is capable of partitioning the user population into preference clusters, and the model is capturing individualized preference structure. 



\begin{figure}[ht]
  \begin{center}
\centerline{\includegraphics[width=0.8\columnwidth]{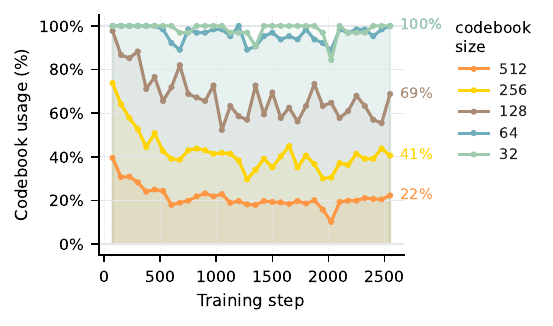}}
    \caption{Codebook utilization rate for different codebook sizes (\textit{window size=30}).}
    \label{fig:co_percent}
  \end{center}
\end{figure}

\textit{Codebook usage analysis.} Figure~\ref{fig:co_percent} reports codebook utilization across training steps for sizes from 32 to 512. We observe two patterns:
(1) Small codebooks ($K \leq 64$) maintain near 100\% utilization throughout training, while larger codebooks stabilize at lower rates. This reflects the codebook's adaptive use of capacity, matching the data's intrinsic preference diversity.
(2) To further illustrate this, the result shows that the absolute number of active entries converges to a similar range across all codebook sizes (see Appendix~\ref{app:co_absolute}).


\section{Conclusion}
We introduced EpiPersona, a novel method for pluralistic alignment that jointly models stable personas and dynamic episodes. By projecting individual preference feedback into a low-dimensional, shared persona space via parameterized abductive reasoning and vector quantization, our method effectively disentangles enduring personal traits from situational variations.



\section*{Limitations}
While EpiPersona demonstrates promising results in open-domain dialogue personalization, several limitations warrant acknowledgment. \textit{First}, our experiments are conducted on open-domain datasets. Future studies could further reflect the complexity of specialized domains. Extending EpiPersona to high-stakes and specialized settings such as medical consultation, legal advisory, and educational contexts would be an important step toward broader impact. \textit{Second}, the datasets employed in this work are predominantly English-based, which restricts our ability to assess the framework's robustness across diverse linguistic contexts. Future work should explore multilingual benchmarks to address this gap.

\section*{Ethical Considerations}

As with any personalization technology, our approach raises inherent \textbf{privacy considerations}. Learned persona representations may unintentionally encode sensitive attributes, such as \textit{political beliefs, health conditions, or socioeconomic status}, or enable inference of private information from behavioral traces, even when no explicit personal data is provided. The episodic memory mechanism, while designed for preference tracking, could in principle reconstruct fine-grained behavioral profiles that users did not intend to disclose.

Although we mitigate this risk by (a) relying solely on publicly available, peer-reviewed benchmarks (Prism and Arena) that have undergone ethics review with established privacy protections, and (b) strictly adhering to dataset usage agreements, the potential for such issues in \textit{real-world deployment} remains a concern. Future work should explore privacy-preserving personalization techniques, such as \textit{differential privacy} over persona embeddings or \textit{federated learning} schemes that keep user-specific parameters on-device.

Finally, while the proposed framework advances preference modeling and reward learning for pluralistic alignment, its application should be approached with caution to prevent \textbf{misuse}. Concretely, a deployed system could (a) exploit fine-grained persona representations to craft persuasive messages tailored to individual psychological vulnerabilities, or (b) repurpose preference signals as a covert profiling mechanism for \textit{surveillance}. To guard against such scenarios, we will release our method under a \textit{restrictive license} that explicitly prohibits commercial exploitation for targeted manipulation or user monitoring. Beyond our own release practices, we call on the community to develop complementary safeguards: (a) fairness evaluation protocols that audit personalization outcomes across demographic groups and episodic contexts, (b) red-teaming methodologies specific to persona-conditioned generation, and (c) institutional best practices for responsible deployment of pluralistic alignment systems.

\bibliography{custom}

@article{kirk2024prism,
  title={The PRISM alignment dataset: What participatory, representative and individualised human feedback reveals about the subjective and multicultural alignment of large language models},
  author={Kirk, Hannah R and Whitefield, Alexander and R{\"o}ttger, Paul and Bean, Andrew and Margatina, Katerina and Ciro, Juan and Mosquera, Rafael and Bartolo, Max and Williams, Adina and He, He and others},
  journal={Advances in Neural Information Processing Systems},
  volume={37},
  pages={105236--105344},
  year={2024}
}

@inproceedings{ryan-etal-2025-synthesizeme,
    title = "{S}ynthesize{M}e! Inducing Persona-Guided Prompts for Personalized Reward Models in {LLM}s",
    author = "Ryan, Michael J.  and
      Shaikh, Omar  and
      Bhagirath, Aditri  and
      Frees, Daniel  and
      Held, William  and
      Yang, Diyi",
    editor = "Che, Wanxiang  and
      Nabende, Joyce  and
      Shutova, Ekaterina  and
      Pilehvar, Mohammad Taher",
    booktitle = "Proceedings of the 63rd Annual Meeting of the Association for Computational Linguistics (Volume 1: Long Papers)",
    month = jul,
    year = "2025",
    address = "Vienna, Austria",
    publisher = "Association for Computational Linguistics",
    url = "https://aclanthology.org/2025.acl-long.397/",
    doi = "10.18653/v1/2025.acl-long.397",
    pages = "8045--8078",
    ISBN = "979-8-89176-251-0"
}

@inproceedings{balepur-etal-2025-whose,
    title = "Whose Boat Does it Float? Improving Personalization in Preference Tuning via Inferred User Personas",
    author = "Balepur, Nishant  and
      Padmakumar, Vishakh  and
      Yang, Fumeng  and
      Feng, Shi  and
      Rudinger, Rachel  and
      Boyd-Graber, Jordan Lee",
    editor = "Che, Wanxiang  and
      Nabende, Joyce  and
      Shutova, Ekaterina  and
      Pilehvar, Mohammad Taher",
    booktitle = "Proceedings of the 63rd Annual Meeting of the Association for Computational Linguistics (Volume 1: Long Papers)",
    month = jul,
    year = "2025",
    address = "Vienna, Austria",
    publisher = "Association for Computational Linguistics",
    url = "https://aclanthology.org/2025.acl-long.168/",
    doi = "10.18653/v1/2025.acl-long.168",
    pages = "3371--3393",
    ISBN = "979-8-89176-251-0"
}

@inproceedings{oh-etal-2025-comparison,
    title = "Comparison-based Active Preference Learning for Multi-dimensional Personalization",
    author = "Oh, Minhyeon  and
      Lee, Seungjoon  and
      Ok, Jungseul",
    editor = "Che, Wanxiang  and
      Nabende, Joyce  and
      Shutova, Ekaterina  and
      Pilehvar, Mohammad Taher",
    booktitle = "Proceedings of the 63rd Annual Meeting of the Association for Computational Linguistics (Volume 1: Long Papers)",
    month = jul,
    year = "2025",
    address = "Vienna, Austria",
    publisher = "Association for Computational Linguistics",
    url = "https://aclanthology.org/2025.acl-long.1590/",
    doi = "10.18653/v1/2025.acl-long.1590",
    pages = "33145--33166",
    ISBN = "979-8-89176-251-0"
}

@inproceedings{li20251,
  title={From 1,000,000 users to every user: Scaling up personalized preference for user-level alignment},
  author={Li, Jia-Nan and Guan, Jian and Wu, Songhao and Wu, Wei and Yan, Rui},
  booktitle={Proceedings of the 64th Annual Meeting of the Association for Computational Linguistics (Volume 1: Long Papers)},
  pages={30144--30168},
  year={2026}
}

@inproceedings{zollo2024personalllm,
 author = {Zollo, Thomas and Siah, Andrew and Ye, Naimeng and Li, Li and Namkoong, Hongseok},
 booktitle = {International Conference on Learning Representations},
 editor = {Y. Yue and A. Garg and N. Peng and F. Sha and R. Yu},
 pages = {66949--66971},
 title = {PersonalLLM: Tailoring LLMs to Individual Preferences},
 url = {https://proceedings.iclr.cc/paper_files/paper/2025/file/a730abbcd6cf4a371ca9545db5922442-Paper-Conference.pdf},
 volume = {2025},
 year = {2025}
}

@inproceedings{zhao2023group,
 author = {Zhao, Siyan and Dang, John and Grover, Aditya},
 booktitle = {International Conference on Learning Representations},
 editor = {B. Kim and Y. Yue and S. Chaudhuri and K. Fragkiadaki and M. Khan and Y. Sun},
 pages = {57965--57987},
 title = {Group Preference Optimization: Few-Shot Alignment of Large Language Models},
 url = {https://proceedings.iclr.cc/paper_files/paper/2024/file/ffd4f5a2ea6b93e9bf5af9264d568cf2-Paper-Conference.pdf},
 volume = {2024},
 year = {2024}
}

@article{poddar2024personalizing,
  title={Personalizing reinforcement learning from human feedback with variational preference learning},
  author={Poddar, Sriyash and Wan, Yanming and Ivison, Hamish and Gupta, Abhishek and Jaques, Natasha},
  journal={Advances in Neural Information Processing Systems},
  volume={37},
  pages={52516--52544},
  year={2024}
}

@article{Askell2021AGL,
  title={A general language assistant as a laboratory for alignment},
  author={Askell, Amanda and Bai, Yuntao and Chen, Anna and Drain, Dawn and Ganguli, Deep and Henighan, Tom and Jones, Andy and Joseph, Nicholas and Mann, Ben and DasSarma, Nova and others},
  journal={arXiv preprint arXiv:2112.00861},
  year={2021}
}

@article{kirk2024benefits,
  title={The benefits, risks and bounds of personalizing the alignment of large language models to individuals},
  author={Kirk, Hannah Rose and Vidgen, Bertie and R{\"o}ttger, Paul and Hale, Scott A},
  journal={Nature Machine Intelligence},
  volume={6},
  number={4},
  pages={383--392},
  year={2024},
  publisher={Nature Publishing Group UK London}
}

@inproceedings{zhang-etal-2025-bridging,
    title = "Bridging Intuitive Associations and Deliberate Recall: Empowering {LLM} Personal Assistant with Graph-Structured Long-term Memory",
    author = "Zhang, Yujie  and
      Yuan, Weikang  and
      Jiang, Zhuoren",
    editor = "Che, Wanxiang  and
      Nabende, Joyce  and
      Shutova, Ekaterina  and
      Pilehvar, Mohammad Taher",
    booktitle = "Findings of the Association for Computational Linguistics: ACL 2025",
    month = jul,
    year = "2025",
    address = "Vienna, Austria",
    publisher = "Association for Computational Linguistics",
    url = "https://aclanthology.org/2025.findings-acl.901/",
    doi = "10.18653/v1/2025.findings-acl.901",
    pages = "17533--17547",
    ISBN = "979-8-89176-256-5"
}

@inproceedings{yuan-etal-2026-lecode,
    title = "{L}e{C}o{D}e: A Benchmark Dataset for Interactive Legal Consultation Dialogue Evaluation",
    author = "Yuan, Weikang  and
      Song, Kaisong  and
      Jiang, Zhuoren  and
      Cao, Junjie  and
      Zhang, Yujie  and
      Lin, Jun  and
      Kuang, Kun  and
      Zhang, Ji  and
      Liu, Xiaozhong",
    editor = "Liakata, Maria  and
      Moreira, Viviane P.  and
      Zhang, Jiajun  and
      Jurgens, David",
    booktitle = "Proceedings of the 64th Annual Meeting of the {A}ssociation for {C}omputational {L}inguistics (Volume 1: Long Papers)",
    month = jul,
    year = "2026",
    address = "San Diego, California, United States",
    publisher = "Association for Computational Linguistics",
    url = "https://aclanthology.org/2026.acl-long.1667/",
    doi = "10.18653/v1/2026.acl-long.1667",
    pages = "36019--36048",
    ISBN = "979-8-89176-390-6"
}

@article{du2025valuesim,
  title={Valuesim: Generating backstories to model individual value systems},
  author={Du, Bangde and Ye, Ziyi and Wu, Zhijing and Monika, Jankowska and Zhu, Shuqi and Ai, Qingyao and Zhou, Yujia and Liu, Yiqun},
  journal={arXiv preprint arXiv:2505.23827},
  year={2025}
}

@article{gu2024survey,
  title={A survey on llm-as-a-judge},
  author={Gu, Jiawei and Jiang, Xuhui and Shi, Zhichao and Tan, Hexiang and Zhai, Xuehao and Xu, Chengjin and Li, Wei and Shen, Yinghan and Ma, Shengjie and Liu, Honghao and others},
  journal={The Innovation},
  volume={7},
  number={6},
  year={2026},
  publisher={Elsevier}
}

@inproceedings{zheng2023judging,
 author = {Zheng, Lianmin and Chiang, Wei-Lin and Sheng, Ying and Zhuang, Siyuan and Wu, Zhanghao and Zhuang, Yonghao and Lin, Zi and Li, Zhuohan and Li, Dacheng and Xing, Eric and Zhang, Hao and Gonzalez, Joseph and Stoica, Ion},
 booktitle = {Advances in Neural Information Processing Systems},
 doi = {10.52202/075280-2020},
 editor = {A. Oh and T. Naumann and A. Globerson and K. Saenko and M. Hardt and S. Levine},
 pages = {46595--46623},
 publisher = {Curran Associates, Inc.},
 title = {Judging LLM-as-a-Judge with MT-Bench and Chatbot Arena},
 url = {https://proceedings.neurips.cc/paper_files/paper/2023/file/91f18a1287b398d378ef22505bf41832-Paper-Datasets_and_Benchmarks.pdf},
 volume = {36},
 year = {2023}
}

@inproceedings{chen2025pal,
      title={{PAL}: Sample-Efficient Personalized Reward Modeling for Pluralistic Alignment},
      author={Chen, Daiwei and Chen, Yi and Rege, Aniket and Wang, Zhi and Vinayak, Ramya Korlakai},
      booktitle={The Thirteenth International Conference on Learning Representations},
      year={2025},
      url={https://openreview.net/forum?id=1kFDrYCuSu}
}

@inproceedings{zhao2023cross,
author = {Zhao, Chuang and Zhao, Hongke and HE, Ming and Zhang, Jian and Fan, Jianping},
title = {Cross-domain recommendation via user interest alignment},
year = {2023},
isbn = {9781450394161},
publisher = {Association for Computing Machinery},
address = {New York, NY, USA},
url = {https://doi.org/10.1145/3543507.3583263},
doi = {10.1145/3543507.3583263},
booktitle = {Proceedings of the ACM Web Conference 2023},
pages = {887–896},
numpages = {10},
location = {Austin, TX, USA},
series = {WWW '23}
}

@inproceedings{zhupersonality,
 author = {Zhu, Minjun and Weng, Yixuan and Yang, Linyi and Zhang, Yue},
 booktitle = {International Conference on Learning Representations},
 editor = {Y. Yue and A. Garg and N. Peng and F. Sha and R. Yu},
 pages = {14206--14255},
 title = {Personality Alignment of Large Language Models},
 url = {https://proceedings.iclr.cc/paper_files/paper/2025/file/25203d1cc8c58381eab578f4fcf9c4f8-Paper-Conference.pdf},
 volume = {2025},
 year = {2025}
}

@inproceedings{guan2025survey,
  title={A survey on personalized Alignment—The missing piece for large language models in real-world applications},
  author={Guan, Jian and Wu, Junfei and Li, Jia-Nan and Cheng, Chuanqi and Wu, Wei},
  booktitle={Findings of the Association for Computational Linguistics: ACL 2025},
  pages={5313--5333},
  year={2025}
}

@inproceedings{crosshypothetical,
  title={Hypothetical minds: Scaffolding theory of mind for multi-agent tasks with large language models},
  author={Cross, Logan and Xiang, Violet and Bhatia, Agam and Yamins, Daniel and Haber, Nick},
  booktitle={International Conference on Learning Representations},
  volume={2025},
  pages={6507--6546},
  year={2025}
}

@inproceedings{xie2024can,
 author = {Xie, Chengxing and Chen, Canyu and Jia, Feiran and Ye, Ziyu and Lai, Shiyang and Shu, Kai and Gu, Jindong and Bibi, Adel and Hu, Ziniu and Jurgens, David and Evans, James and Torr, Philip and Ghanem, Bernard and Li, Guohao},
 booktitle = {Advances in Neural Information Processing Systems},
 doi = {10.52202/079017-0501},
 editor = {A. Globerson and L. Mackey and D. Belgrave and A. Fan and U. Paquet and J. Tomczak and C. Zhang},
 pages = {15674--15729},
 publisher = {Curran Associates, Inc.},
 title = {Can Large Language Model Agents Simulate Human Trust Behavior?},
 url = {https://proceedings.neurips.cc/paper_files/paper/2024/file/1cb57fcf7ff3f6d37eebae5becc9ea6d-Paper-Conference.pdf},
 volume = {37},
 year = {2024}
}

@inproceedings{riemerposition,
  author       = {Matthew Riemer and
                  Zahra Ashktorab and
                  Djallel Bouneffouf and
                  Payel Das and
                  Miao Liu and
                  Justin D. Weisz and
                  Murray Campbell},
  title        = {Position: Theory of Mind Benchmarks are Broken for Large Language
                  Models},
  booktitle    = {Forty-second International Conference on Machine Learning, {ICML}
                  2025, Vancouver, BC, Canada, July 13-19, 2025 - Position Paper Track},
  publisher    = {OpenReview.net},
  year         = {2025},
  url          = {https://proceedings.mlr.press/v267/riemer25a.html},
  bibsource    = {dblp computer science bibliography, https://dblp.org}
}

@misc{jang2023personalized,
      title={Personalized Soups: Personalized Large Language Model Alignment via Post-hoc Parameter Merging}, 
      author={Joel Jang and Seungone Kim and Bill Yuchen Lin and Yizhong Wang and Jack Hessel and Luke Zettlemoyer and Hannaneh Hajishirzi and Yejin Choi and Prithviraj Ammanabrolu},
      year={2023},
      eprint={2310.11564},
      archivePrefix={arXiv},
      primaryClass={cs.CL},
      url={https://arxiv.org/abs/2310.11564}, 
}

@article{bradley1952rank,
  title={Rank analysis of incomplete block designs: I. the method of paired comparisons},
  author={Bradley, Ralph Allan and Terry, Milton E},
  journal={Biometrika},
  volume={39},
  number={3/4},
  pages={324--345},
  year={1952},
  publisher={JSTOR}
}

@inproceedings{xu2025personalized,
    title = "Personalized Generation In Large Model Era: A Survey",
    author = "Xu, Yiyan  and
      Zhang, Jinghao  and
      Salemi, Alireza  and
      Hu, Xinting  and
      Wang, Wenjie  and
      Feng, Fuli  and
      Zamani, Hamed  and
      He, Xiangnan  and
      Chua, Tat-Seng",
    editor = "Che, Wanxiang  and
      Nabende, Joyce  and
      Shutova, Ekaterina  and
      Pilehvar, Mohammad Taher",
    booktitle = "Proceedings of the 63rd Annual Meeting of the Association for Computational Linguistics (Volume 1: Long Papers)",
    month = jul,
    year = "2025",
    address = "Vienna, Austria",
    publisher = "Association for Computational Linguistics",
    url = "https://aclanthology.org/2025.acl-long.1201/",
    doi = "10.18653/v1/2025.acl-long.1201",
    pages = "24607--24649",
    ISBN = "979-8-89176-251-0"
}

@article{zhang2024personalization,
  title={Personalization of large language models: A survey},
  author={Zhang, Zhehao and Rossi, Ryan A and Kveton, Branislav and Shao, Yijia and Yang, Diyi and Zamani, Hamed and Dernoncourt, Franck and Barrow, Joe and Yu, Tong and Kim, Sungchul and others},
  journal={arXiv preprint arXiv:2411.00027},
  year={2024}
}

@article{radford2018improving,
  title={Improving language understanding by generative pre-training},
  author={Radford, Alec and Narasimhan, Karthik and Salimans, Tim and Sutskever, Ilya and others},
  year={2018},
  publisher={San Francisco, CA, USA}
}

@inproceedings{cai2026one,
  author = {Cai, Hongru and Li, Yongqi and Yu, Tiezheng and Zhu, Fengbin and Wang, Wenjie and Feng, Fuli and Li, Wenjie},
  title = {One Adapts to Any: Meta Reward Modeling for Personalized LLM Alignment},
  year = {2026},
  isbn = {9798400725999},
  publisher = {Association for Computing Machinery},
  address = {New York, NY, USA},
  url = {https://doi.org/10.1145/3805712.3809689},
  doi = {10.1145/3805712.3809689},
  pages = {110--121},
  numpages = {12},
  location = {Australia},
  series = {SIGIR '26},
  booktitle = {Proceedings of the 49th International ACM SIGIR Conference on Research and Development in Information Retrieval}
}

@article{boselore,
  title={Lore: Personalizing llms via low-rank reward modeling},
  author={Bose, Avinandan and Xiong, Zhihan and Chi, Yuejie and Du, Simon Shaolei and Xiao, Lin and Fazel, Maryam},
  journal={arXiv preprint arXiv:2504.14439},
  year={2025}
}

@misc{bai2022training,
      title={Training a Helpful and Harmless Assistant with Reinforcement Learning from Human Feedback}, 
      author={Yuntao Bai and Andy Jones and Kamal Ndousse and Amanda Askell and Anna Chen and Nova DasSarma and Dawn Drain and Stanislav Fort and Deep Ganguli and Tom Henighan and Nicholas Joseph and Saurav Kadavath and Jackson Kernion and Tom Conerly and Sheer El-Showk and Nelson Elhage and Zac Hatfield-Dodds and Danny Hernandez and Tristan Hume and Scott Johnston and Shauna Kravec and Liane Lovitt and Neel Nanda and Catherine Olsson and Dario Amodei and Tom Brown and Jack Clark and Sam McCandlish and Chris Olah and Ben Mann and Jared Kaplan},
      year={2022},
      eprint={2204.05862},
      archivePrefix={arXiv},
      primaryClass={cs.CL},
      url={https://arxiv.org/abs/2204.05862}, 
}

@article{christiano2017deep,
  title={Deep reinforcement learning from human preferences},
  author={Christiano, Paul F and Leike, Jan and Brown, Tom and Martic, Miljan and Legg, Shane and Amodei, Dario},
  journal={Advances in neural information processing systems},
  volume={30},
  year={2017}
}

@article{kaufmann2025survey,
  author       = {Timo Kaufmann and
                  Paul Weng and
                  Viktor Bengs and
                  Eyke H{\"{u}}llermeier},
  title        = {A Survey of Reinforcement Learning from Human Feedback},
  journal      = {Trans. Mach. Learn. Res.},
  volume       = {2025},
  year         = {2025},
  url          = {https://openreview.net/forum?id=f7OkIurx4b},
  bibsource    = {dblp computer science bibliography, https://dblp.org}
}

@inproceedings{Zhu2024AddressingRC,
    title   = {Addressing Representation Collapse in Vector Quantized Models with One Linear Layer},
    author  = {Yongxin Zhu and Bocheng Li and Yifei Xin and Linli Xu},
    year    = {2024},
    url     = {https://api.semanticscholar.org/CorpusID:273812459}
}

@article{van2017neural,
  title={Neural discrete representation learning},
  author={Van Den Oord, Aaron and Vinyals, Oriol and others},
  journal={Advances in neural information processing systems},
  volume={30},
  year={2017}
}

@article{bengio2013estimating,
  title={Estimating or propagating gradients through stochastic neurons for conditional computation},
  author={Bengio, Yoshua and L{\'e}onard, Nicholas and Courville, Aaron},
  journal={arXiv preprint arXiv:1308.3432},
  year={2013}
}

\appendix

\setcounter{table}{0}   
\setcounter{figure}{0}
\setcounter{section}{0}
\setcounter{equation}{0}

\renewcommand\thetable{A.\arabic{table}}
\renewcommand\thefigure{A.\arabic{figure}}
\renewcommand\theequation{A.\arabic{equation}}

\renewcommand*{\theHtable}{\thetable}
\renewcommand*{\theHfigure}{\thefigure}
\renewcommand*{\theHequation}{\theequation}
\renewcommand*{\theHsection}{\thesection}

\section{Appendix}
\subsection{Task design}
\label{app:task}

These two settings define the \textit{general evaluation paradigm} rather than being tied to any specific model design.

\paragraph{Task 1: Generative setting.}

In this paradigm, an LLM is prompted to infer a natural-language description of an individual user’s preference based on historical interactions $X_u$, optionally conditioned on the current episode $e$.
This formulation is applicable to a broad class of \emph{non-parametric} baselines that rely on in-context prompting.

\begin{equation}
\label{eq:judge_task}
\text{Pref} = G(X_u, e)\ \ or \ \ \text{Pref} = G(X_u)
\end{equation}

The generated preference $\text{Pref}$ is expressed in natural language and summarizes the individual’s persona or preference.
Subsequently, preference and a set of candidate responses $\{\tau_e^{(1)}, \tau_e^{(2)}\}$ are provided to a large language model acting as a judge. $\hat{y}_e \in \{0,1\}$ indicates whether the user selects the response.
The judge predicts which response the user is more likely to select:
\begin{equation}
\hat{y}_e
=
LLM(\text{Pref}, \tau_e^{(1)}, \tau_e^{(2)}).
\end{equation}

This task is mainly designed for \textbf{EpiPersona-A} and non-parametric preference modeling, following the LLM-as-a-judge evaluation paradigm.

\paragraph{Task 2: Discriminative setting.}
Let $e$ denote the current episode.
The user’s historical preference information is $X_u$.
Given a candidate response $\tau_e$ under episode $e$, the goal is to predict the probability that the user would select this response:



\begin{equation}
\hat{y}_e
=
\arg\max_{\tau_e \in \{\tau_e^{(1)}, \tau_e^{(2)}\}}
\; \mathbb{P}(y_e = 1 \mid h(X_u), \tau_e).
\end{equation}
where $h$ denotes the encoding of user context $X_u$. Model performance is evaluated using accuracy (Acc).


This task is primarily designed for \textbf{EpiPersona-B} and parametric preference modeling.

These two task settings correspond to different real-world application scenarios. The generative setting is better suited for plug-and-play settings, where the generated textual preference descriptions can be directly injected into the prompt of the LLM. The discriminative setting, on the other hand, is more appropriate for scenarios that require end-to-end training of personalized reward models. The two variants operate in fundamentally different output spaces, thereby providing a more comprehensive evaluation of the EpiPersona framework as a whole.

    
    

\subsection{Dataset construction}
\label{app:dataset}
A fundamental challenge in personalized preference research is that collecting real-world preference data with individual identity annotations is exceedingly difficult. Arena and Prism are among the few publicly available datasets that simultaneously satisfy the following criteria:
(1) Each dataset contains multiple preference records per user across different episodes, enabling genuine longitudinal modeling of individual preferences.
(2) Both datasets exhibit sufficient inter-individual heterogeneity to support meaningful pluralistic alignment research, capturing the diversity of real-world user populations.

\begin{figure}
   \begin{center}
    \centerline{\includegraphics[width=0.8\linewidth]{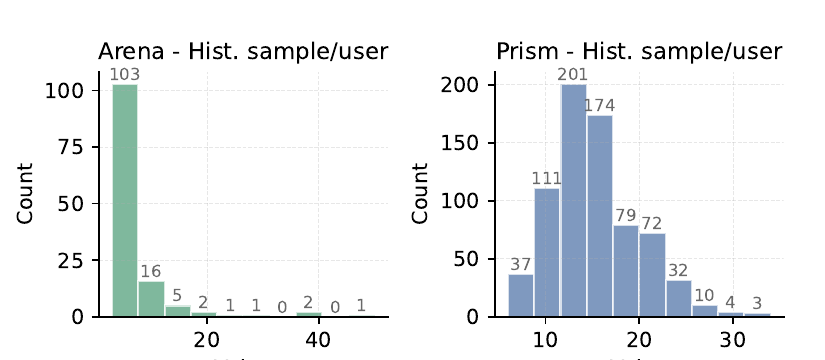}}
    \caption{The distribution of the history samples per user.}
    \label{fig:data_dist}
\end{center}
\end{figure}

We follow the existing study \cite{ryan-etal-2025-synthesizeme} in adopting the dataset split (train, validation, and test sets) as well as the division of each individual’s historical and current data to ensure more consistent and continuous experimental validation. These two datasets compile the high-quality and challenging individual data to serve as a benchmark for personalized alignment. The distribution of historical sample numbers per individual is shown in Figure \ref{fig:data_dist}. The statistics of historical/current pairs for both datasets are summarized in Table~\ref{tab:data_stats}.



\begin{table}[htbp]
\caption{Statistics of user numbers and historical/current pairs for Arena and Prism datasets.}
\label{tab:data_stats}
\centering
\scalebox{0.9}{
\begin{tabular}{lcc}
\toprule
 & Arena & Prism \\
\cmidrule(lr){2-2} \cmidrule(lr){3-3}
 & Hist/cur pairs & Hist/cur pairs \\
\midrule
Train & 202/103 & 4192/2177 \\
Validation & 155/80 & 973/514 \\
Test & 514/\textbf{284} & 5816/\textbf{3033} \\
\bottomrule
\end{tabular}
}
\end{table}

\subsection{Implementation details}
\label{app:detail}
The model is fine-tuned with LoRA (rank $r=8$, $\alpha=32$, dropout $=0.1$). The vector quantization module uses a codebook size of 256 for Prism and 64 for Arena. The VQ commitment loss weight is set to 1.0, the diversity loss weight to 0.1, and the diversity softmax temperature to 0.5. User preference history is encoded with a sliding window of size 30. Preference descriptions $s$ are generated by Qwen3-235B-A22B-Instruct based on users' real feedback. We use a JSON dictionary format with fields such as \textit{persona}, \textit{value}, \textit{preferred style}, and \textit{intent} (see prompt in Appendix \ref{box:guidance}.1).

Training is conducted on 2 NVIDIA GPUs (H20) with a per-device batch size of 2 and gradient accumulation steps of 2, yielding an effective global batch size of 8. We use the AdamW optimizer with learning rate $5\times10^{-6}$, weight decay $0.01$, and a linear warmup over 100 steps followed by constant   
scheduling. Gradient clipping is applied at 1.0. Distributed training is managed via DeepSpeed ZeRO Stage 2 with CPU optimizer offloading. Due to the limited size of the Arena dataset, we first pre-trained the transformer-based model on the Prism dataset (for EpiPersona-A) to stabilize training. We initialize the codebook via K-means rather than random initialization, with 20 iterations at initialization. We reset a codebook entry when its EMA usage frequency falls below 1.


\subsection{Comparison of preference modeling methods across core capabilities}
\label{app:acompare}
We formalize the core capabilities of preference modeling as: human-observable and interpretable (Human-O/I), machine-understandable (Machine-U), compression capability (Compact-C), and global preference abstraction capability (Global-A).

\textit{Text-based summarization} and RAG offer interpretability but provide weaker compression, learnability, and abstraction, which may undesirably focus on users’ fragmented information. \textit{Parameterized preference} modeling excels in learnability and abstraction but is less observable and plug-and-play. \textit{Predefined dimensions} compress preferences into dense vectors, aiding learning and interpretability, yet rely heavily on manual design, limiting flexibility.

Our approach proposes a learnable personalized preference attribution (EpiPersona-A) that capitalizes on the strengths of these four capabilities. Specifically, \textit{Machine-understandable:} EpiPersona-A parameterizes users’ static personas employing parameterized abductive reasoning and VQ-based mapping; \textit{Compression capability:} Parameterized static personas serve as a highly compressed representation of user preferences, preventing loss of persona information via text summarization; \textit{Global preference abstraction capability:} EpiPersona-A provides a macro-level, abstract characterization of user personalities, instead of isolated or fragmented local information; \textit{Human-observable and interpretable:} Dynamic preferences are predicted in textual form, making them plug-and-play compatible with any LLM. See Table \ref{tab:preference_models} for the comparison of these models.

\begin{table*}[h]
\caption{Comparing various preference modeling methods in terms of key attributes. 
Predef.: Predefined dimensions; 
Text-based sum.: text-based preference summarization; 
Parameterized.: Parameterized modeling; 
EpiPersona-A: Our proposed method.}
\label{tab:preference_models}
\centering
\scalebox{0.85}{%
\begin{tabular}{lcccc}
\toprule
\textbf{Method} & \textbf{Human-O/I} & \textbf{Machine-U} & \textbf{Compact-C} & \textbf{Global-A} \\
\midrule
Text-based sum. & \Checkmark & \textbf{-}  & \textbf{-}  & \textbf{-}  \\
Parameterized. & \textbf{-}  & \Checkmark & \Checkmark & \Checkmark \\
Predef. & \Checkmark & \Checkmark & \Checkmark & \textbf{-}  \\
RAG & \Checkmark & \textbf{-}  & \textbf{-}  & \textbf{-}  \\
\midrule
EpiPersona-A & \Checkmark & \Checkmark & \Checkmark & \Checkmark \\
\bottomrule
\end{tabular}%
}
\end{table*}


\subsection{Figure of model details}
\begin{figure*}
    \begin{center}
\centerline{\includegraphics[width=0.8\textwidth]{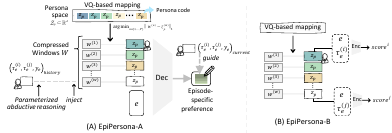}}
    \caption{Model architecture details (parameterized abductive reasoning and VQ-based mapping).}
    \label{fig:detail_epipersona}
  \end{center}
\end{figure*}

Figure \ref{fig:detail_epipersona} illustrates the detailed architecture of our model, including the parameterized abductive reasoning module and the VQ-based mapping mechanism.

\subsection{The detail of VQ-based mapping}
\label{app:ema_commit}
\paragraph{Exponential moving average update.} 
We adopt an exponential moving average scheme to stabilize codebook learning.
For each persona codeword $z_p^{(m)}$, we maintain an EMA count $N_m$ and an EMA embedding $E_m$. At each update step, we first accumulate statistics over the current batch:
\begin{align}
N_m &\leftarrow \gamma N_m + (1 - \gamma)
\sum_{w=1}^{n} \mathbb{I}\!\left[m^{*(w)} = m\right], \\
E_m &\leftarrow \gamma E_m + (1 - \gamma)
\sum_{w=1}^{n} \mathbb{I}\!\left[m^{*(w)} = m\right] \nonumber\\
&\quad {} \times \textbf{w}^{(w)}
\end{align}
where $\gamma \in (0,1)$ is the EMA decay rate and $\mathbb{I}[\cdot]$ denotes the indicator function. The persona codeword is then updated as:
\begin{equation}
z_p^{(m)} \leftarrow \frac{E_m}{N_m}.
\end{equation}



\begin{figure*}[ht]
  \begin{center}
\centerline{\includegraphics[width=0.8\textwidth]{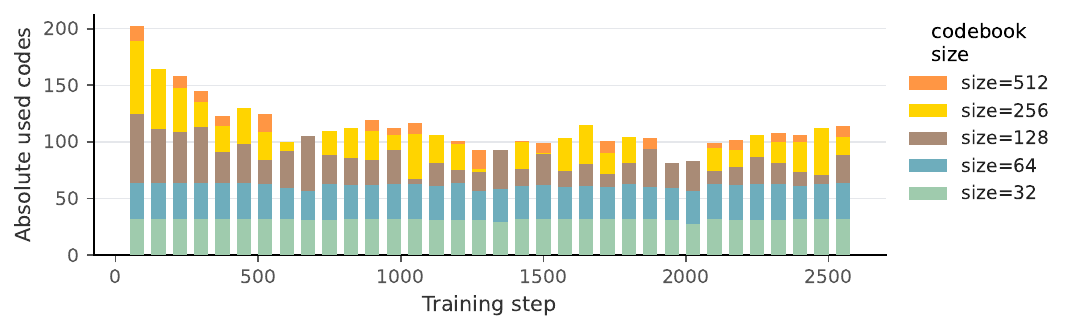}}
    \caption{Codebook utilization across different codebook sizes during training (\textit{window size=30}).}
    \label{fig:co_absolute}
  \end{center}

\end{figure*}
\begin{figure*}[ht]
  \begin{center}
\centerline{\includegraphics[width=0.8\textwidth]{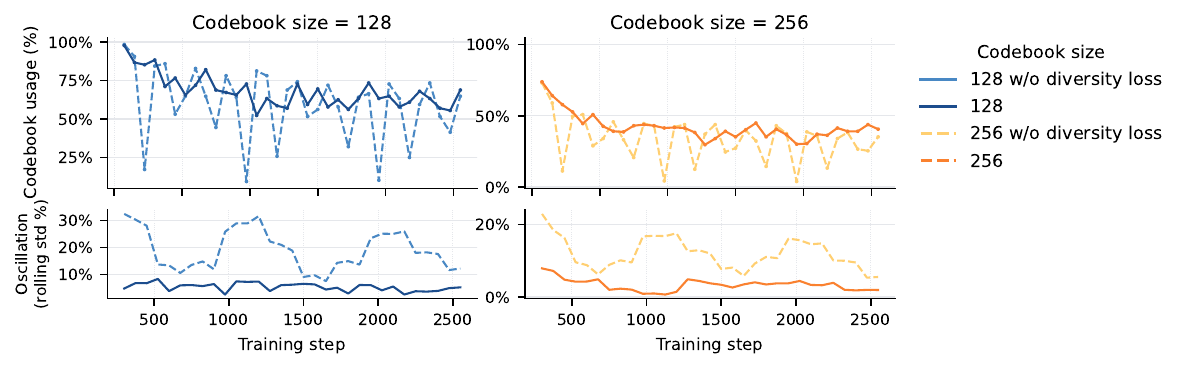}}
    \caption{Ablation study on the diversity loss. Top: codebook utilization rate over training steps. Bottom: oscillation measured by rolling standard deviation. The diversity loss reduces both codebook collapse and utilization instability across codebook sizes 128 and 256.}
    \label{fig:diversity_ablation}
  \end{center}

\end{figure*}
\begin{figure*}[ht]
  \begin{center}
\centerline{\includegraphics[width=0.8\textwidth]{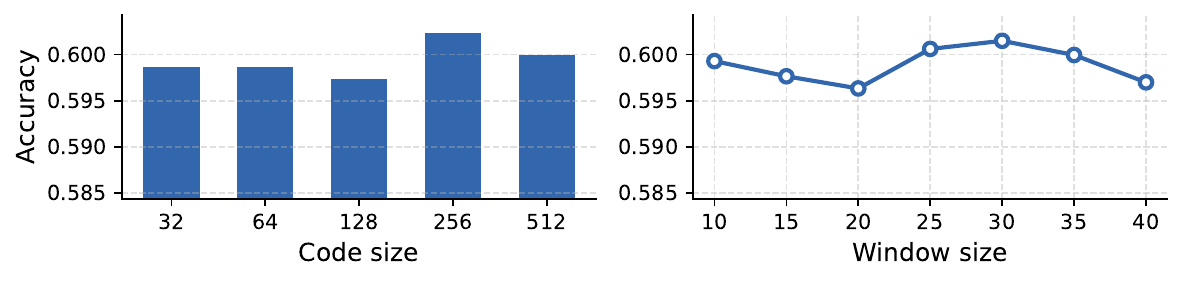}}
    \caption{Hyperparameter experiment on \textit{Prism}.}
    \label{fig:hyper}
  \end{center}
\end{figure*}


\subsection{Codebook usage analysis}
\label{app:co_absolute}
As shown in Figure~\ref{fig:co_absolute}, the codebook usage dynamics reveal several consistent patterns across all codebook sizes:

\begin{enumerate}

    \item \textbf{Codebook utilization dynamics.} At the very beginning of training, codebook usage starts high due to the $K$-means initialization. After the initialization phase, the actual number of utilized codewords gradually stabilizes as training progresses.

    \item \textbf{Size-dependent utilization ceiling.} Larger codebooks (size $\in \{256, 512\}$) converge to high absolute utilization levels, whereas smaller codebooks (size $\in \{32, 64\}$) saturate earlier with nearly all entries occupied. This indicates that an excessively small codebook reaches its utilization ceiling prematurely, thereby limiting the model's expressive power.

    \item \textbf{Dataset-driven convergence.} As training progresses, codebooks of varying sizes converge to their respective stable usage levels. This implies that the effective number of required codebook entries is closely tied to the intrinsic preference diversity of the dataset, rather than being an artifact of model capacity alone.
\end{enumerate}

\subsection{Effect of diversity loss on codebook utilization}
\label{app:diversity_ablation}

To validate the effectiveness of the diversity loss, we conduct an ablation study comparing codebook usage with and without it under two codebook sizes (128 and 256), as shown in Figure~\ref{fig:diversity_ablation}. To quantify 
training stability, we measure oscillation as the rolling standard deviation of codebook utilization over a window of 4 consecutive training steps.

\paragraph{Reduced Oscillation.}
Without the diversity loss, codebook usage exhibits severe oscillation throughout training. The rolling standard deviation reaches up to 30\% for codebook size 128 and approximately 20\% for codebook size 256. With the 
diversity loss, oscillation is suppressed to below 10\% across both settings, indicating substantially more stable training dynamics.

\paragraph{Alleviated Codebook Collapse.}
Without the diversity loss, codebook utilization frequently drops below 25\% for size 128 and below 10\% for size 256, reflecting recurring collapse. With the diversity loss, utilization is maintained consistently above 50\% and 30\% for the two sizes, respectively. These results confirm that the diversity loss is essential for stable and efficient codebook utilization.

\subsection{Hyperparameter experiment}
\label{app:hyper}
To investigate the effect of key hyperparameters, we conduct ablation studies on codebook size and window size, as shown in Figure \ref{fig:hyper}.

For codebook size, accuracy remains stable across {32, 64, 128, 256, 512}. For window size, performance follows an inverted U-shaped trend over {10, 15, 20, 25, 30, 35, 40}. A window that is too small fails to capture sufficient context, while an overly large window introduces irrelevant noise. In both cases, accuracy variance remains within 0.5\%, indicating that our model is relatively robust to these hyperparameters.









\subsection{Inference algorithm}
We summarize the inference procedure for EpiPersona. Given user preference feedback and an episode, the algorithms first encode feedback using parameterized abductive reasoning and convert it to latent persona representations via VQ-based mapping. We then generate either episode-specific preferences (EpiPersona-A) or reward scores for candidate responses (EpiPersona-B) (see algorithm \ref{alg:epipersona-a} and \ref{alg:epipersona-b-mod}).


\begin{algorithm}[tb]
  \caption{EpiPersona-A: Persona-episode coupled preference generation}
  \label{alg:epipersona-a}
  \begin{algorithmic}
    \STATE {\bfseries Input:} Feedback set $X_u$ for individual $u$; target episode $e = (\tau_e^{(i)}, \tau_e^{(j)})$
    \STATE {\bfseries Output:} Personalized preference description $\text{Pref}_e$
    \STATE \textit{// Step 1: Parameterized abductive reasoning (Section~\ref{sec:abd})}
    \FOR{each feedback $x = (\tau_e^{(i)}, \tau_e^{(j)}, y_e) \in X_u$}
      \STATE Infer estimated persona: $r \leftarrow \lambda(\mathcal{P},\, \tau_e^{(i)},\, \tau_e^{(j)},\, y_e)$
    \ENDFOR
    \STATE Collect all estimated personas: $R_u = \{r \mid x \in X_u\}$
    \STATE \textit{// Step 2: VQ-based mapping (Section~\ref{sec:vq})}
    \STATE Project into shared space: $\mathbf{w}^{(w)} = f_\theta(r)$ for each $r \in R_u$
    \STATE Quantize each window via nearest-neighbor lookup:
    \[
      m^{*(w)} = \arg\min_{m \in \{1,\dots,P\}} \|\mathbf{w}^{(w)} - z_p^{(m)}\|_2
    \]
    \STATE Obtain latent persona set: $Z_u = \bigl(z_p^{(m^{*(w)})}\bigr)_{w=1}^{n}$
    \STATE \textit{// Step 3: Persona-episode coupling (Section~\ref{sec:coupling})}
    \STATE Generate preference: $\text{Pref}_e = g_\psi(Z_u,\, e)$
    \STATE Format $\text{Pref}_e$ as JSON with keys: \texttt{persona}, \texttt{value}, \texttt{identification}, \texttt{intent}, \texttt{style}
    \STATE \textbf{return} $\text{Pref}_e$
  \end{algorithmic}
\end{algorithm}


\begin{algorithm}[tb]
  \caption{EpiPersona-B: Episode-aware persona-based reward modeling}
  \label{alg:epipersona-b-mod}
  \begin{algorithmic}
    \STATE {\bfseries Input:} Feedback set $X_u$ for individual $u$; episode $e$; candidate responses $\tau_e^{(i)}, \tau_e^{(j)}$
    \STATE {\bfseries Output:} Rewards $R_\theta(Z_u, \tau_e^{(i)})$ and $R_\theta(Z_u, \tau_e^{(j)})$
    \STATE \textit{// Step 1: Parameterized abductive reasoning (Section~\ref{sec:abd})}
    \FOR{each feedback $x = (\tau_e^{(i)}, \tau_e^{(j)}, y_e) \in X_u$}
      \STATE Infer estimated persona: $r \leftarrow \lambda(\mathcal{P},\, \tau_e^{(i)},\, \tau_e^{(j)},\, y_e)$
    \ENDFOR
    \STATE Collect all estimated personas: $R_u = \{ r \mid x \in X_u \}$
    \STATE \textit{// Step 2: VQ-based mapping (Section~\ref{sec:vq})}
    \STATE Project into shared space: $\mathbf{w}^{(w)} = f_\theta(r)$ for each $r \in R_u$
    \STATE Quantize each window via nearest-neighbor lookup:
    \[
      m^{*(w)} = \arg\min_{m \in \{1,\dots,P\}} \|\mathbf{w}^{(w)} - z_p^{(m)}\|_2
    \]
    \STATE Obtain latent persona set: $Z_u = \bigl(z_p^{(m^{*(w)})}\bigr)_{w=1}^{n}$
    \STATE \textit{// Step 3: Persona-episode coupling \& reward computation (Section~\ref{sec:coupling})}
    \FOR{each candidate response $\tau_e^{(k)} \in \{\tau_e^{(i)}, \tau_e^{(j)}\}$}
      \STATE Compute episode-persona coupled representation: $h_\theta(Z_u,\, \tau_e^{(k)})$
      \STATE Compute reward: $R_\theta(Z_u,\, \tau_e^{(k)}) = \mathbf{W}^\top \phi\bigl(h_\theta(Z_u,\, \tau_e^{(k)})\bigr)$
    \ENDFOR
    \STATE Predict preference via Bradley-Terry comparison:
    \[
      \hat{y}_e = \mathbf{1}\!\left[ R_\theta(Z_u, \tau_e^{(i)}) > R_\theta(Z_u, \tau_e^{(j)}) \right]
    \]
    \STATE \textbf{return} $R_\theta(Z_u, \tau_e^{(i)})$, $R_\theta(Z_u, \tau_e^{(j)})$, $\hat{y}_e$
  \end{algorithmic}
\end{algorithm}

\clearpage


\subsection{Prompt}
This section lists the prompts used for modeling user beliefs and preferences, including:  \\
1. Prompt for fine-grained preference description extraction;\\  
2. LLM-as-a-Judge Prompt for EpiPersona-A;  \\
3. LLM-as-a-Judge Prompt with Retrieval Augmentation;  \\
4. Text-based Episode-specific Preference Prediction Prompt \textit{(used in ablation test)};  \\
5. Parameterized Abductive Reasoning Prompt ($\mathcal{P}$);\\  
6. EpiPersona-A Prompt for episode-specific context coupled with user representation $Z_u$.

\begin{tcolorbox}[
  title={1. Prompt for fine-grained preference description extraction (guidance for EpiPersona-A)},
  colback=white,
  colframe=blue!70!black,
  breakable
]
\label{box:guidance}
\small
\begin{Verbatim}[breaklines=true, breaksymbolleft={}, breakanywhere=true]
Given the dialogue context below and two possible responses—
one selected by the user and one not:

Dialogue context: {context}

Chosen answer: {chosen}

Rejected answer: {rejected}

---Task---
Analyze the user's selection and extract explicitly defined
preference elements.

---Definitions---
The extracted preferences should be divided into the following
categories:

1. Stable preferences (long-term, consistent traits):
persona: the user's personality or character traits.
value: the user's values, opinions, or attitudes.
identification: the user's social identity, role, or profession
if inferable.

2. Dynamic preferences (context-dependent and situational):
style: the user's preferred expression style, tone, or format.
intent: the user's goal or intention in choosing the response.

---Rules---
1. If a preference element cannot be inferred, return None.
2. Do not guess or invent information beyond what can be
   reasonably inferred from the context.
3. The output must be strictly formatted as JSON and must not
   contain comments.

---Output Format---
{
  "stable_preferences": {
    "persona": "<persona>",
    "value": "<value>",
    "identification": "<identification>"
  },
  "dynamic_preferences": {
    "style": "<style>",
    "intent": "<intent>"
  }
}
\end{Verbatim}
\normalsize
\end{tcolorbox}

\begin{tcolorbox}[
  title={2. LLM-as-a-Judge Prompt (EpiPersona-A)},
  colback=white,
  colframe=blue!70!black
]
\small
\begin{Verbatim}[breaklines=true, breaksymbolleft={}, breakanywhere=false]
Your input fields are:
1. conversation (str): The conversation context leading up to the completions.
2. first_completion (str): The first of the two candidate responses.
3. second_completion (str): The second of the two candidate responses.

Your output fields are:
1. reasoning (str): A brief explanation of the judgment.
2. preference (str): The response more likely to be preferred by the user.
   The value must be one of: "First" or "Second".

All interactions follow the structure below.

[[ ## conversation ## ]]
{conversation}

[[ ## first_completion ## ]]
{first_completion}

[[ ## second_completion ## ]]
{second_completion}

The output must be a valid JSON object with the following structure:

{
  "reasoning": "<reasoning>",
  "preference": "<First | Second>"
}

Your objective:
Given a conversation and two responses generated by different models,
determine which response the human user is more likely to prefer.

The user you are judging for has the following preference profile:
===
{user_preference}
===

Consider the user's persona and preferences carefully, and
make a rational, preference-consistent judgment.
\end{Verbatim}
\normalsize
\end{tcolorbox}

\begin{tcolorbox}[
  title={3. LLM-as-a-Judge Prompt with Retrieval Augmentation},
  colback=white,
  colframe=blue!70!black
]
\small
\begin{Verbatim}[breaklines=true, breaksymbolleft={}, breakanywhere=false]
User History (Retrieved Demonstrations):

---
User's historical selection i (can be used as a reference if relevant):

[[ ## conversation ## ]]
{demo_conversation_i}

[[ ## first_completion ## ]]
{demo_first_completion_i}

[[ ## second_completion ## ]]
{demo_second_completion_i}

User's choice: "First" or "Second"

(The above block may repeat for multiple retrieved demonstrations.)

Current Dialogue Context:
{conversation}

Candidate Responses:

[[ ## first_completion ## ]]
{first_completion}

[[ ## second_completion ## ]]
{second_completion}

Task:
Given the current dialogue context and the user's historical selections,
infer which candidate response the user is more likely to choose.
You should support your judgment by reasoning about how the current
responses relate to patterns observed in the user's history.

Output Format:
The output must be a valid JSON object and contain no additional text.

{
  "preference": "<First | Second>"
}
\end{Verbatim}
\normalsize
\end{tcolorbox}

\begin{tcolorbox}[
  title={4. Prompt for Text-based Preference Modeling (Episode-specific Preference Prediction)},
  colback=white,
  colframe=blue!70!black
]
\small
\begin{Verbatim}[breaklines=true, breaksymbolleft={}, breakanywhere=false]
Given the user's historical stable preferences and the current context,
infer the Stable preferences and Dynamic preferences that are relevant
to the specific episode.

Historical Stable Preferences:
{stable_preferences}

Current Context:
{context}

Definitions:

1. Stable preferences (long-term, consistent traits):
- persona: the user's personality or character traits.
- value: the user's values, opinions, or attitudes.
- identification: the user's social identity, role, or profession
  (if inferable).

2. Dynamic preferences (context-dependent and situational):
- style: the user's preferred style, tone, or format of expression.
- intent: the user's immediate goal or purpose in this context.

Rules:
- Infer only preferences that are reasonably supported by the
  historical data and the current context.
- If a preference element cannot be inferred, set its value to null.
- Do not introduce unsupported assumptions.
- The output must be valid JSON and contain no extra text.

Output Format:
{
  "related_stable_preferences": {
    "persona": "<persona>",
    "value": "<value>",
    "identification": "<identification>"
  },
  "dynamic_preferences": {
    "style": "<style>",
    "intent": "<intent>"
  }
}
\end{Verbatim}
\normalsize
\end{tcolorbox}

\begin{tcolorbox}[
  title={5. Parameterized Abductive Reasoning ($\mathcal{P}$)},
  colback=white,
  colframe=blue!70!black
]
\small
\begin{Verbatim}[breaklines=true, breaksymbolleft={}, breakanywhere=false]
Task:
Given a dialogue context and two alternative responses—one chosen by
the user and one rejected—identify the stable preference implied by
the user's selection.

Definition of Stable Preference:
A stable preference reflects enduring aspects of the user's persona,
values, or identity. Such preferences are transferable across dialogue
contexts and are unlikely to change due to short-term or situational
factors.

Inputs:

Dialogue Context:
{context}

Chosen Answer:
{chosen}

Rejected Answer:
{rejected}

Inference Objective:
Infer the stable preference that best explains why the chosen answer
was preferred over the rejected one. If no stable preference can be
reasonably inferred, return an empty string.

Output Format:
The output must be a valid JSON object and contain no additional text.

{
  "stable_preferences": "<preference or empty string>"
}
\end{Verbatim}
\normalsize
\end{tcolorbox}

\clearpage

\begin{tcolorbox}[
  title={6. EpiPersona-A Prompt for Episode-specific Context (Coupled with $Z_u$)},
  colback=white,
  colframe=blue!70!black
]
\small
\begin{Verbatim}[breaklines=true, breaksymbolleft={}, breakanywhere=false]
Given the current context, infer Stable preferences and Dynamic
preferences relevant to the specific context.

Context:
{context}

Respond ONLY in the following JSON structure:

{
    "related_stable_preferences": {
        "persona": "<persona>",
        "value": "<value>",
        "identification": "<identification>"
    },
    "dynamic_preferences": {
        "style": "<style>",
        "intent": "<intent>"
    }
}
\end{Verbatim}
\normalsize
\end{tcolorbox}

\end{document}